# Training Autonomous Drones for Search and Rescue with Convolutional Autoencoder-Generated Wind Simulations

Jiahao Wu, Yang Ye, Jing Du, Ph.D.

**ABSTRACT**

Drones are becoming indispensable in emergency search and rescue (SAR), particularly in intricate urban areas where rapid and accurate response is crucial. This study addresses the pressing need for enhancing drone navigation in such complex, dynamic urban environments, where obstacles like building layouts and varying wind conditions create unique challenges. Particularly, the need for adapting drone's autonomous navigation in correspondence with dynamic wind conditions in urban settings is emphasized, as it is important for drones to avoid loss of control or crashes during SAR. This paper introduces a pioneering method integrating multi-objective reinforcement learning (MORL) with a convolutional autoencoder to train autonomous drones in comprehending and reacting to aerodynamic features in urban SAR. MORL enables the drone to optimize multiple goals, while the convolutional autoencoder generates synthetic wind simulations with a substantially lower computation cost compared to traditional computational fluid dynamics (CFD) simulations. A unique data transfer structure is also proposed, which fosters a seamless integration of perception and decision-making between Machine Learning (ML) and Reinforcement Learning (RL) components. This approach uses imagery data, specific to building layouts, allowing the drone to autonomously formulate policies, prioritize navigation decisions, optimize paths, and mitigate the impact of wind, all while negating the necessity for conventional aerodynamic force sensors. The method was validated with a model of New York City, offering substantial implications for enhancing automation algorithms in urban SAR. This innovation enables the possibility of more efficient, precise, and timely drone SAR operations within intricate urban landscapes.

# 1. INTRODUCTION

Unmanned Aerial Vehicles (UAVs) or drones have gained significant popularity due to their efficiency, versatility, and flexibility in various mission spaces. One of these growing applications is to support search and rescue (SAR) given drones' capacity to navigate challenging disaster-stricken environments [1, 2]. Especially, the advances of autonomous drones equipped with advanced control systems and sensors, such as LiDAR and thermal imaging sensors, have revolutionized SAR efforts [3]. Autonomous drones can efficiently collect data while in flight, supporting tasks like rapid damage assessment, survivor detection, and situational awareness in disaster-affected areas [4-6]. Moreover, autonomous drones are capable of planning flight paths without human intervention, allowing them to adapt to dynamic environmental conditions that are often encountered during disaster response efforts [7]. In the chaotic aftermath of disasters, the ability to reach remote or hazardous locations quickly is crucial for effective response. Autonomous drones can navigate through debris, unstable terrain, and challenging weather conditions, enabling them to provide critical assistance where it is needed most [8]. This autonomy also plays a vital role in relieving human responders from repetitive and potentially hazardous tasks, ultimately improving overall efficiency and safety during disaster search and rescue operations [9].

Central to the implementation of autonomous drones in missions is effective path planning. This involves determining the best route for a drone from its origin to its destination, factoring in requirements like obstacle avoidance and minimizing travel time [10]. While prevailing path-planning approaches prioritize identifying the shortest route and evading obstacles [11], real-world scenarios introduce unpredictable factors, such as changing weather conditions, that can impact the navigation of autonomous drones. For instance, urban settings characterized by tall buildings can disrupt aerodynamics, affecting the intended trajectory of drone flight. Similarly, unexpected turbulence amid closely spaced structures can destabilize drones [12]. On a larger scale, the arrangement of buildings in regions like metropolitan areas intricately influences wind patterns, rendering preset drone path plans less effective [13]. Widespread dynamic winds, including strong gusts, can compromise drone stability and control, while even mild winds can alter flight paths. Additionally, headwinds challenge drone battery life and flight duration, demanding higher power consumption to counter wind resistance. These dynamic environmental factors collectively undermine the conventional path planning of autonomous drones, impeding their ability to safely and efficiently achieve goals.

This paper presents a novel approach to training drone path planning using multi-objective reinforcement learning (MORL). The primary objective is to enable drones to navigate mild wind zones and avoid strong wind zones safely during SAR. The training aims to empower autonomous drones to learn effective strategies for adapting to varying aerodynamic conditions while accomplishing navigation tasks through a reinforcement learning network. The challenge pertains to the creation of a training environment that reproduces the aerodynamic features of a large urban area. Traditionally, this has been done with computational fluid dynamics (CFD). Nevertheless, CFD simulations usually take significant amount of computing resources[14]. Combined with the need for executing MORL, the training of autonomous drones for reacting or avoiding wind zones in addition to target reaching and obstacle avoidance goals would be impractical to finish. As a result, we use a convolutional neural network (CNN)-based model to generate wind data solutions aligned with computational fluid dynamics (CFD) principles. This approach significantly reduces the time required to create wind data compared to conventional methods like the Navier-Stokes equations solver, such as OpenFOAM.

Another unique feature of our approach is the use of sparse sensor. Our system utilizes an onboard camera as the sole sensor, alongside GPS, to facilitate environmental perception and error-based learning for autonomous drones. In other words, no aerodynamic sensors or environmental data about the wind zones will be utilized. We expect this to reflect the real-world scenarios of drone systems in SAR. We posit that there is a latent relationship between building layouts and the wind zone distribution, thus MORL is expected to derive effective wind-reacting strategies from visual information. To obtain visual information efficiently, we adopt curriculum learning during the MORL training process, progressively exposing the model to more intricate and challenging examples to enhance its learning efficiency and effectiveness. Upon completion of the training, the autonomous drone should adeptly perceive various environmental and physical conditions in its proximity. It will then formulate dynamic policies to prioritize navigational decisions, optimizing its path while mitigating adverse environmental consequences. This paper introduces our study's foundation, the proposed method and system structure, case studies of drone training, and a validation case in Manhattan, New York City.

## 2. RELATED WORKS

### 2.1. Autonomous Drones

Autonomous drones have garnered substantial attention owing to their capacity to function without direct human intervention [7]. The adaptability of these autonomous drones has led to a diverse array of applications spanning various industries, including agriculture, disaster management, infrastructure inspection, and wildlife monitoring [15] [16] [17] [18]. These autonomous drones employ diverse self-navigation techniques, encompassing GPS-based, vision-based, and sensor-based methods for path planning and flight control. Among these, GPS-based navigation stands out as the most prevalent and well-established method, relying on GPS data to navigate along predetermined waypoints [19]. Access to a comprehensive global map that provides a high-level perspective of the mission area empowers the drone to plan its route and navigate obstacles with heightened efficiency. [20]. Even in partially unfamiliar and dynamic 3D environments, access to GPS data can facilitate hybrid navigation approaches, empowering autonomous drone operations, such as swift responses to newly detected obstacles [21].

However, numerous drone missions are conducted in GPS-denied environments characterized by towering buildings or dense vegetation, impeding the use of GPS-based navigation [22]. Consequently, alternatives like vision-based and sensor-based techniques are being explored. Vision-based navigation relies on computer vision algorithms and onboard cameras to perceive and analyze the surroundings, enabling drones to circumvent obstacles and pursue designated paths [23]. Illustrative examples of vision-based navigation systems encompass Simultaneous Localization and Mapping (SLAM) [24] and Visual-Inertial Odometry (VIO) [25]. Sensor-based navigation strategies harness sensors such as LiDAR, radar, and ultrasonic devices to gauge distances between the drone and adjacent objects. This proves valuable in environments with restricted or compromised visual navigational capabilities [26]. These techniques frequently integrate sensor data, amalgamating information from diverse sources to enhance the accuracy and resilience of the navigation system [27]. More recently, machine learning integration has markedly bolstered autonomous drones' capabilities in intricate environments [28]. Notably, deep learning methods, including Convolutional Neural Networks (CNNs) [29] and Recurrent Neural Networks (RNNs), have been deployed to refine vision-based navigation, empowering superior object recognition, segmentation, and tracking [30].

Adopting LiDAR-based sensors is a prominent solution regarding sensing techniques for autonomous drones. El-Sheimy and Li [31] introduced an innovative algorithm to minimize computational overhead and energy usage for indoor LiDAR target detection. Furthermore, camera-based methodologies offer a more intricate visual depiction of the surroundings, enabling drones to respond to various obstacles. Chakravarty et al. [32] employed a single forward-facing camera and trained a CNN for depth perception, guiding the drone's obstacle avoidance. Sani and Karimian [33] proposed a comprehensive solution for indoor drone navigation employing two cameras, a Kalman filter, and an inertial sensor. García Carrillo et al. [34] affixed a stereoscopic camera to a drone. They trained a CNN to process depth maps, successfully detecting and locating drones up to eight meters away with a 10% error margin.

The literature has actively explored advanced algorithms capable of effectively processing supplementary sensor data across diverse conditions [35]. Sun et al. [36] introduced the Rapidly-exploring Random Tree (RRT) algorithm for intricate environment path planning. It has gained prominence in UAV path planning due to its swift search for space exploration and the generation of viable paths. Wen et al.[37] subsequently presented an improved iteration of RRT, termed Safe-RRT, ensuring both safety and optimality in UAV paths within cluttered settings. Optimization-centric methodologies have also gained attention for addressing path planning complexities. Albert et al. [38] utilized mixed-integer linear programming (MILP) to simultaneously create collision-free paths for multiple UAVs, showcasing its prowess in managing intricate scenarios. Similarly, Huang et al.[39] delved into particle swarm optimization (PSO) for real-time drone path planning, underscoring its efficacy in optimal path generation.

## 2.2. Reinforcement Learning for Self-Navigation with Limited Data

Reinforcement learning (RL) constitutes a subset of machine learning techniques wherein an agent is trained to make decisions through interactions with its environment [40]. This process involves the agent maximizing its cumulative reward by learning from trial-and-error experiences, thus uncovering an optimal policy for goal achievement [41]. Notably, RL differentiates itself by lacking a predetermined training dataset, relying instead on continuous feedback to adapt actions and optimize final rewards. This adaptability is a crucial advantage for autonomous drone navigation in dynamic environments. Noteworthy implementations of RL in autonomous drone navigation include the work of Pham et al. [42], who employed a PID+Q-learning strategy to enable the drone to navigate unfamiliar obstacle-free environments. Hodge et al.[43]developed a Unity-based 2D training environment to safely train drones in obstacle avoidance using LiDAR sensor data and Proximal Policy Optimization. Muñoz et al. [44] utilized DDQN and a custom neural network named JNN to navigate 3D simulated environments, integrating depth camera data for obstacle recognition. Multi-Objective Reinforcement Learning (MORL) addresses complex decision-making with competing objectives. It generates diverse solutions, forming a Pareto frontier to accommodate conflicting goals. Ramezani Dooraki and Lee[45] constructed a MORL network to train a drone navigation system in dynamic settings, encompassing stationary and mobile obstacles. Shantia et al. [46] introduced a two-stage training approach for drone navigation based on visual input, revealing the superior performance of the multi-objective training network compared to the single-objective counterpart

In summary, Multi-Objective Reinforcement Learning presents a well-established framework related to multi-dimensional optimization challenges, making it a proper choice for optimizing drone navigation while considering wind dynamics within urban landscapes. The inclusion of MORL gives us the tools to reach the complex balance between different factors, providing a range of solutions that cater to various preferences of decision-makers. The success of MORL training for SAR drones relies on the ability to comprehend diverse environment conditions. In this study, we focus on wind distributions. It requires a simulation environment that can reproduce the real-world complexity of aerodynamic features, especially in the presence of building structures in urban setting. Existing methods heavily rely on CFD, which uses numerical analysis and data structures to analyze and solve problems that involve fluid flows [14]. Although accurate in simulation results, CFD usually takes significant amounts of computing time to accomplish. A method for generating high-quality wind simulation data is needed.

**2.3. Synthetic Aerodynamic Data Generation**

In autonomous drone training, the simulation of wind conditions has emerged as a pivotal component for enhancing the navigational capabilities of these unmanned aerial vehicles. Recent research endeavors have extensively explored wind simulation methods, offering valuable insights and tools to equip drones with the skills required to operate effectively in real-world environments. Various approaches have been employed, ranging from computational fluid dynamics (CFD) simulations to physics-information neural networks simulations (PINN). Computational simulations utilizing CFD models allow for a comprehensive analysis of wind patterns and their impact on drone flight. Paz, et al. [47] utilized Computational Fluid Dynamics (CFD) simulations to replicate the airflow produced by the propellers of a quadcopter Unmanned Aerial Vehicle (UAV). They investigated how aerodynamic factors influenced the stability of flight when the quadcopter was operating near walls and the ground. Qu, et al. [48] used CFD model to generate local wind data for reinforcement learning training in multi-drone coordination of disaster response work. Giersch, et al. [49] created a new Large Eddy Simulation (LES) model to simulate the turbulent flow around the urban building structures and proposed its potential in the application of drone operations. Jeong, et al. [50] employed a Large Eddy Simulation to replicate the wind patterns influenced by urban terrain. They utilized output from a Weather Research and Forecasting (WRF) model as the initial and boundary conditions for LES, creating an accurate and complex urban wind environment for drone training. While CFD or LES simulations offer a versatile means of subjecting drones to various wind scenarios, their computational cost is prohibitively high, making it challenging to apply them to reinforcement learning tasks that demand substantial amounts of data input. To avoid this cost problem, a simpler wind methodology was used in the drone simulation environment. Chu, et al. [51] used sUAS flight simulation software to analyze how wind speed, direction, and turbulence affect sUAS performance in following planned flight paths and sustaining missions. They manually created multiple stable winds to test the drone's behavior under different conditions. However, this methodology has limitations in replicating all the intricate features of wind distribution due to its oversimplification. To enhance the efficiency and quality of wind data acquisition, deep learning technology is integrated with traditional simulation methods. Physics-Information Neural Networks (PINN) is an innovative approach at the intersection of deep learning and computational fluid dynamics (CFD). It has gained significant attention in recent years for its ability to efficiently model and simulate fluid flow phenomena while respecting the underlying physics governing fluid behavior. Vuppala and Kara [52] introduced a time-efficient machine-learning approach aimed at generating realistic wind data. This method ensures the safe operation of small unmanned aircraft systems (sUASs) in urban environments by predicting future flow patterns using existing Large Eddy Simulation (LES) data. Milla-Val, et al. [53] introduced an approach for obtaining wind prediction results similar to those generated by computational fluid dynamics (CFD). These results were less detailed, but more cost-effective, which was accomplished through the application of supervised learning techniques. Ma, et al. [54] explored using physics-driven Convolutional Neural Networks (CNN) to predict complex, high-resolution flow fields. The numerical experiments demonstrated that this approach achieved first-order accuracy, accelerated training for multiple cases, and accurately predicted flow patterns. By integrating these diverse wind simulation methods into autonomous drone training protocols, researchers and engineers are actively contributing to the advancement of drone technology, enabling these unmanned systems to navigate safely and efficiently in the face of challenging wind conditions.

## 3. METHODOLOGY

### 3.1. Overview

This research introduces an innovative reinforcement learning framework for training drones to achieve multivariate objectives, or MORL. These objectives include the drone's autonomous navigation within urban landscapes and obstacle avoidance in a dynamic wind environment, all while successfully reaching specified targets. To be noted, in order to ensure the efficiency of the learning process, we address the need for simulating aerodynamic features due to wind distribution in urban setting with a physics-informed neural network DeepCFD. Unlike the common practice of utilizing CFD for aerodynamic simulation, DeepCFD would enable the generation of synthetic wind distribution data with satisfactory accuracy [55]. This model generates wind map data by leveraging the architectural layout of obstructive buildings. The system adeptly captures the nuanced and intricate interactions between the fabricated surroundings and the complex patterns of varying wind zones throughout the learning process. In addition, in contrast to previous studies that assumed drones possessed prior access to global information or were equipped with an array of sensors for comprehensive environmental data acquisition, our approach assumes that the drone is equipped solely with limited imagery sensors like RGB cameras. **Fig. 1** shows the comprehensive structure of the proposed MORL framework.

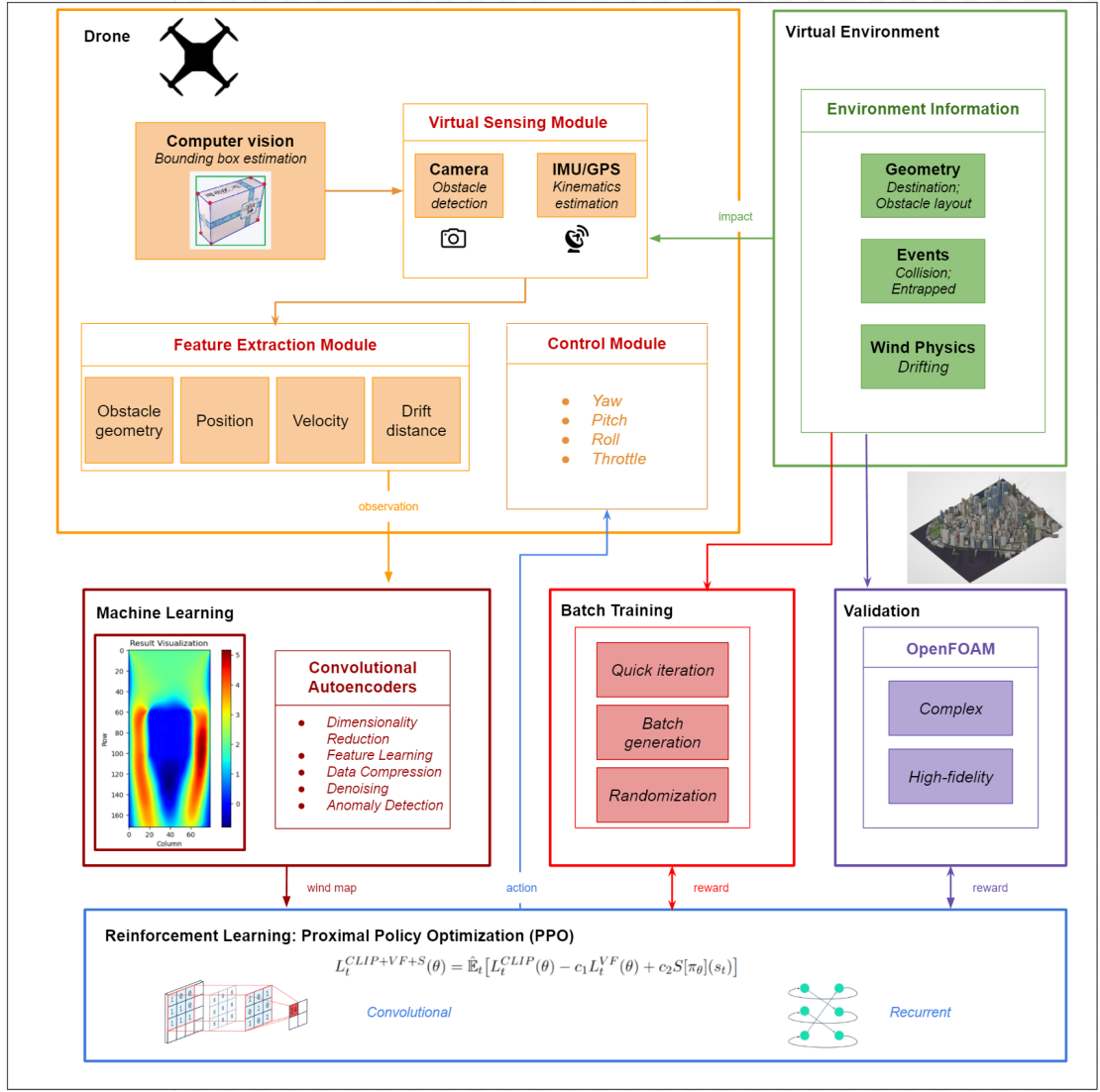

**Fig. 1** The framework of the proposed MORL method

As shown in **Fig.1**, the drone agent exclusively takes in processed camera data as external environmental inputs. We employ a 3D bounding box algorithm to process the visual data derived from the front depth camera, encompassing imagery of constructed environments, notably buildings, into 3D vectors. The whole training progress can be divided into two successive phases. During the initial phase, training unfolds within a simulation devoid of wind zones. This is achieved by employing the Proximal Policy Optimization (PPO) algorithm [56]and an LSTM structure[57]. The PPO algorithm, recognized for its stability, efficiency, sample efficacy, and robustness, is an apt choice for training drones to cultivate proficient policies with a relatively modest number of samples[43]. Besides, the LSTM model is instrumental in retaining information from preceding time steps within a sequence. The recurrent nature of this model facilitates the capture of temporal dependencies within the input data, enhancing the drone's decision-making process.

Our traininr takes a two-step process. The first step primarily focuses on building the feasibility of the RL architecture, constructing the navigation system, and bolstering the drone's competence in obstacle avoidance. During the second step, the training framework integrates dynamics related to wind influence. This entails the inclusion of aerodynamic forces within simulated environments, particularly when the drone approaches buildings. Given the multi-objective nature of MORL in our study, which includes target achievement, obstacle avoidance, and mitigation of detrimental effects caused by dynamic wind zones on the drone's trajectory and stability, a multivariate reward function is used. This function facilitates the drone's learning to evade potent wind areas, react to milder wind conditions, circumvent visible obstacles, and devise alternative routes to the target whenever necessary. After the training phases, real-world data is introduced to validate the attained results. We use OpenFOAM [58], an open-source computational fluid dynamics (CFD) software, to generate accurate CFD results for a selected metropolitan district within New York City for validation.

## 3.2. Data Transfer Structure

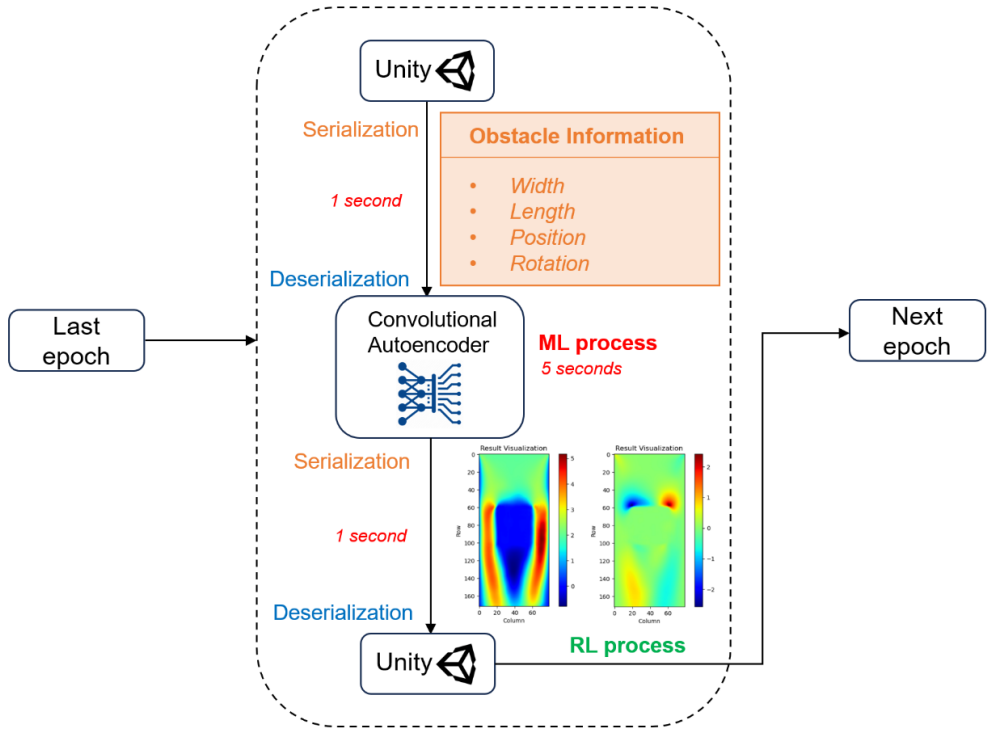

(a). Sequential data transfer framework

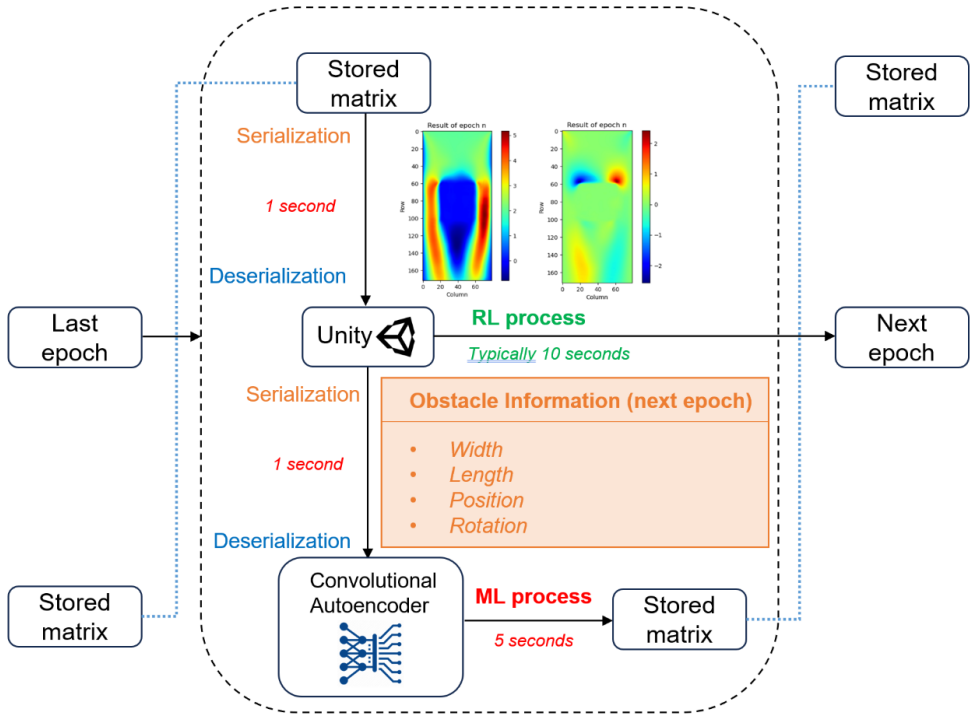

(b). Parallel data transfer framework

**Fig. 2** The comparison between two analytical pipelines

A challenge in MORL training is the required efficiency to ensure enough number of epochs. Although the traditional CFD software is used to expedite the generation of synthetic wind distribution data, we found that the data generation and training speed were less efficient as expected due to the complexity of the urban layout and wind distribution features. As a result, we propose a novel data transfer and analytical pipeline for allowing parallel processing of data generation and RL training within the same epoch. **Fig. 2** illustrates a comparative analysis between two distinct flowchart approaches employed during the training phase. This study introduces an innovative data transfer structure designed to expedite the training process. The machine learning model typically takes around five seconds to generate CFD results based on obstacle information. As shown in **Fig. 2(a),** in the conventional sequential data transfer framework, obstacle information layout for a given epoch is initially generated within the Unity simulation environment. This layout is then transmitted to Python for machine-learning computations. Subsequently, the results are returned to Unity, serving as wind map information for reinforcement learning. However, due to the serial nature of this data transfer mechanism, the reinforcement learning phase commences only upon receiving machine learning results, resulting in significant time delays throughout extensive training iterations. To enhance the overall training speed, we have restructured the data transfer framework between the machine learning model and reinforcement training, as shown in **Fig. 2(b)**, rendering them parallel processes. Specifically, our new data transfer structure employs two memory spaces within Unity to store obstacle layout information for both the current and upcoming epochs rather than just the current one. Additionally, Python utilizes a memory space to retain machine learning model outcomes. During each epoch, Unity transmits the layout information for the forthcoming epoch to the Python machine-learning model, which promptly responds with results from the previous interaction stored in memory. Unity also possesses another memory space to store obstacle layout information linked to CFD results from the prior machine-learning cycle. This arrangement enables the reinforcement learning process to initiate without waiting for machine learning computations, effectively eradicating delays stemming from data transfer across distinct platforms. Moreover, CFD results for the next epoch are computed concurrently with reinforcement learning training. Consequently, if the time required for reinforcement learning in one epoch exceeds that of the machine learning process, there is no surplus time delay in the overall training regimen. This optimization significantly curtails time wastage attributed to data transfer between disparate platforms. In our test it was confirmed that our new data transfer and analytical pipelibe could reduce the time delay by about 80%, from 7 seconds to 1 second for each training epoch.

### 3.3. Convolutional Autoencoder for Synthetic Aerodynamic Data Generation

We have developed a comprehensive methodology to generate and simulate the aerodynamic characteristics of the wind around tall buildings. Although CFD has conventionally used to simulate aerodynamic forces, our investigation has revealed its inefficiency in meeting the demands of extensive simulations requisite for the Deep Reinforcement Learning (DRL) algorithm. The necessities of DRL typically entail executing thousands of simulated scenarios marked by dynamic conditions. The computational cost of subjecting each scenario to CFD simulations proves notably substantial. As such, we propose a CNN encode-decode architecture to generate synthetic CFD results with an accurate rate that can reach 98% (**Fig.3**). This architecture takes the geometry information, and does the encode process. After the encoder, it decodes all the features and then simulates CFD results.

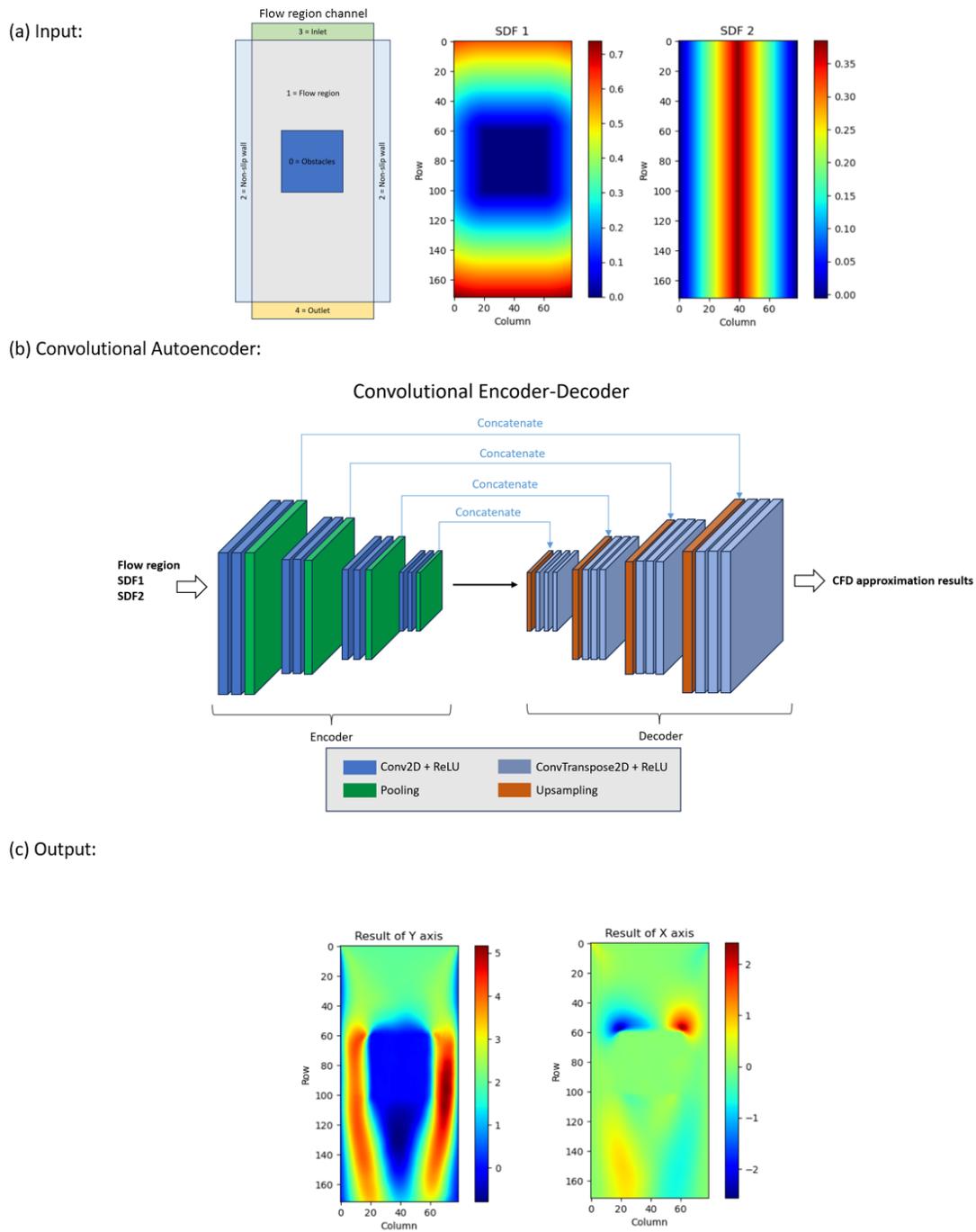

**Fig. 3** The convolutional autoencoder for CFD synthetic data generation.

A particular challenge to the Physics-Informed Neural Networks (PINN) method for synthtic CFD data generation is the so-called Navier - Stokes equations. Navier - Stokes equations refer to a set of equations that describe the motion of fluid substances. These equations are based on the principles of conservation of momentum, and they define how the velocity of fluid changes over time given the forces acting upon it. For CFD problems, it means that the boundary conditions and the location of the obstacle can significantly impact the final results. Therefore, the input of the CNN must contain information related to the geometry and the obstacle position. To present the obstacle and boundary information to the CNN model, we developed a method of parameterization to deliver the information. Specifically, we define five conditions using numbers from 0 to 5. As shown in **Fig.3 (a)**, the length of the flow region is 172, and the width is 79. In the flow region, 0 corresponds to the obstacle, 1 signifies the free-flow area, 2 denotes the upper/bottom no-slip wall condition, 3 represents the constant velocity inlet condition, and 4 indicates the zero-gradient velocity outlet condition. Then, the CNN model can know the specific geometry for each case and incorporate it as an input. In addition, we applied the Signed Distance Function (SDF) to represent the spatial information among the flow region. A Signed Distance Function (SDF) is a mathematical representation used in computer graphics, computational geometry, and physics simulations to describe the distance between a point in space and a geometric shape (typically in 2D or 3D). The unique characteristic of an SDF is that it can represent both the distance and the direction from a point to the closest point on the shape's surface. Additionally, it encodes whether the point is inside or outside the shape by using a positive or negative sign, respectively. To better represent the geometry distance information, we use two signed distance functions as the input of the proposed CNN model.

Mathematically, in our case, for a point x in space X and a geometric shape S, the first signed distance function SDF(x) is defined as:

$$SDF_1(x) = \begin{cases} -d(x, \partial S) & if\ x \in S \\ d(x, \partial S) & else \end{cases} \qquad Eq(1)$$

Where $\partial S$ refers to the boundary of the geometric shape S, which is the obstacle in our case. $d(a, b)$ is the function of finding the minimal spatial distance between point a and shape b. The results are negative for any points in the geometric shape S, which helps the CNN model identify the specific spatial relationship between the points and the obstacles.

The second signed distance function SDF(x) is defined as:

$$SDF_2(x) = \begin{cases} -L(x, \theta) & if\ x \in S \\ L(x, \theta) & else \end{cases} \quad Eq(2)$$

Where $\theta$ refers to the center of the obstacle. $L(x, \theta)$ is the function of calculating the horizontal distance between the certain point the obstacle center.

One of the cores in this machine learning model is the Convolutional Neural Network (CNN), depicted in **Fig.3 (b)**. In recent years, CNNs have achieved groundbreaking results in various domains, such as computer vision, medical imaging, and even autonomous driving [59]. Their success can be attributed to their remarkable ability to automatically extract relevant features from raw data, drastically reducing the need for manual feature engineering. What sets CNNs apart from traditional neural networks is their ability to automatically learn hierarchical patterns and features directly from the raw pixel data. This characteristic makes CNNs especially well-suited for tasks where spatial relationships and local patterns play a crucial role in decision-making. In our project, the wind distribution has a strong relationship with the position of the obstacles, which is very suitable for the CNN model. Another core of our synthetic wind data generation is the Autoencoders (AEs). An AE is an artificial neural network used in machine learning for unsupervised learning tasks, particularly in dimensionality reduction, feature learning, and data compression [60]. The encoder compresses the input data into a lower-dimensional representation or latent space. This process involves transforming the original data, which might be high-dimensional and complex, into a more compact and abstract representation. The decoder takes the low-dimensional representation (the encoding) generated by the encoder and reconstructs an approximation of its original input data. This network can learn an efficient representation of the input data while minimizing the reconstruction error, which is designed for feature learning.

This combination leverages the strengths of both CNNs (which are good at learning features from grid-like data) and AEs (which excel at compressing and reconstructing data). The goal is to create a specialized model that extracts features from images and then rebuilds them effectively. As shown in **Fig.3 (c)**, the model's output gives us predicted results for Computational Fluid Dynamics (CFD), using input details like the shape, obstacles, and distances. An error function (MAE) measures the gap between the model's output and the actual CFD results during training. This comparative evaluation informs adjustments to the network's weights, tuning its predictive capacity. With the application of this Convolutional Autoencoder, we can generate CFD approximation results effectively compared to the traditional Navier - Stokes equations solvers.

## 3.4. Deep Reinforcement Learning Network

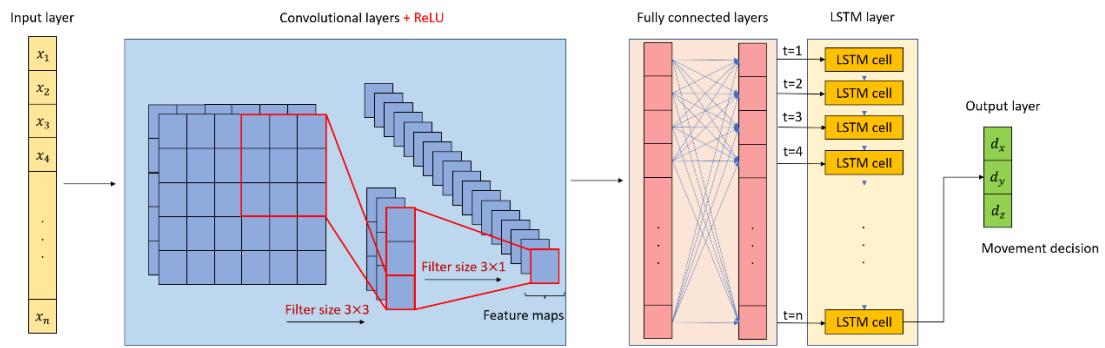

**Fig. 4** The architecture of the deep reinforcement learning network

The fundamental concept behind the MORL technique lies in a deep reinforcement learning (DRL) framework, which empowers the drone to navigate through dynamic environments adeptly. Our approach involves the development of a DRL network comprising seven distinct layers, as depicted in **Fig. 4**. Commencing with the initial layer, which consists of 87 inputs, 72 of these inputs are dedicated to capturing and retaining local geometric data. In our scenario, this pertains to using bounding box results obtained from the drone's onboard imagery sensor, representing obstacles, primarily buildings, situated within the drone's visual proximity. For each obstacle encountered, the 3D bounding box algorithm facilitates the derivation of all eight spatial positions of its vertices. These spatial positions encompass three values apiece, corresponding to the x, y, and z axes. This innovative approach enables the encapsulation of crucial geometric attributes of a building within a concise array. The remaining 12 inputs document comprehensive information concerning the drone agent, such as the *drone-to-target distance*, the *present drone velocity*, the *drone's location*, and the *drone's drifting* due to the wind force. Notably, each input assumes the form of a vector comprising three distinct values. To strengthen the drone's capability to circumvent strong wind areas, another three inputs have been allocated to retain the memory of the most recent intense wind zone. This information subsequently contributes to the reward function.

After the input layer, there are two convolutional layers. Given the interdependent nature of vertex data in bounding boxes, the integration of convolutional layers proves instrumental in extracting pivotal features. Subsequently, two concealed layers, housing 256 nodes each, follow suit. These layers facilitate the network's capacity to capture intricate interrelationships embedded within the input characteristics. An individual LSTM layer follows the concealed layers, effectively harnessing its recurrent architecture to capture protracted dependencies and temporal patterns in sequential data. This feature proves indispensable in evading perpetual loops during navigation. The ultimate layer in the network materializes as the output layer, generating three outputs. These outputs precisely denote the directional movements the drone should execute across the x, y, and z axes.

In general, its decision network can be defined as Eq (3):

$$D_t = F(b_t, d_t, v_t, w_t, p_t, D_{t-1}) \qquad \text{Eq (3)}$$

Where $D_t$ denotes the decision regarding the drone's movement for the present step, $b_t$ represents the obstacle information derived from the 3D bounding box algorithm. This encompasses data related to the obstacle's relative position and dimensions, $d_t$ represents the relative distances spanning all directions between the drone agent and its designated target, $v_t$ represents the current velocity of the drone, $w_t$ is the wind-related data that stems from the drone's movement and subsequent drift, $p_t$ refers to the local position of the drone, which is used to regulate the drone's flight within a predefined operational range, $D_{t-1}$ is the movement choice made during the prior step.

## 3.5. Reward Function

The core objective of MORL is for the agent is to grasp a sequence of actions that will lead to the highest overall rewards as time passes. These rewards stem from accomplishing various smaller goals step by step while the agent navigates its surroundings. After each action, the agent gets feedback in the form of rewards. The context of this reward scheme involves the status of the drone, environmental factors, and obstacles information. The details of the reward function used in this system are presented in Eq (4).

$$R = \beta_1 r_{distance} + \beta_1 r_{target} + \beta_2 r_{wind} + \beta_3 r_{collision} + \beta_4 r_{stuck} + \beta_5 r_{timeout} + \beta_6 r_{time} \quad \text{Eq (4)}$$

Where $R$ embodies the collective reward available to the drone, this aggregate is formed by blending a selection of distinct reward components: $r_{distance}$ is a changeable reward based on the proximity of the drone to the target location, $r_{target}$ is a binary reward received when the drone successfully reaches its destination without incidents, $r_{wind}$ represents the reward associated with navigating through challenging wind conditions, $r_{collision}$ represents the cumulative penalty as the drone collides with obstacles during its flight, $r_{stuck}$ is a penalty applied when the drone becomes trapped in an infinite navigation loop, $r_{time}$ is a penalty term based on the total duration of the navigation task, and $r_{timeout}$ is a binary term that applies a penalty when the drone exhausts its allotted time without completing the task. $\beta_i$ refers to importance weight, where $\sum_{i=1}^{m} \beta_i = 1$. For simplicity, we kept the weighting of these individual sub-goals uniform in this current study. However, the dynamics of how varying priorities among these sub-goals could influence the overall performance of the MORL in drone navigation warrant further investigation. Subsequent sections delve into deeper insights concerning each of these individual reward components.

The initial reward component motivates the drone to approach the designated target. This continuous reward exerts its influence throughout each step within an epoch. Notably, our experimental framework involves altering the drone and target positions at the onset of each epoch. To ensure uniformity in the distance reward across all epochs, we have formulated the distance reward function as delineated in Eq (5).

$$r_{distance} = \sum_{1}^{T} \frac{dis_t - dis_{t-1}}{dis_0} \times 100 \qquad \text{Eq (5)}$$

Where $r_{distance}$ embodies the potential rewards attainable by the drone, $dis_t$ is the current distance between the drone and the target, $dis_{t-1}$ is the previous distance calculated at the last step, $dis_0$ serves as a constant reference denoting the initial drone-to-target distance at the onset of the present epoch.

The next reward factor offers a positive reward when the drone reaches the target quickly and safely. This reward is given when the drone effectively avoids strong winds and obstacles. As described in Eq (6), this approach encourages the drone to learn how to navigate well in areas with strong winds.

$$r_{target} = \begin{cases} +10 & \text{if arrives target} \\ 0 & \text{otherwise} \end{cases} \qquad \text{Eq (6)}$$

The third reward prompts the drone to explore alternative paths in order to dodge areas with higher wind intensity. In each epoch, if the drone successfully navigates around a region with strong winds, it will be granted a favorable reward, as outlined in Eq (7).

$$r_{wind} = \begin{cases} +10 & \text{if avoids strong wind zone} \\ 0 & \text{otherwise} \end{cases} \qquad \text{Eq (7)}$$

The fourth component of the reward function is a penalty term designed to discourage the drone from engaging in collisions. In real scenarios, collisions with obstacles can yield catastrophic consequences for the drone's operation. To avoid this kind of crash, if a collision occurs in training, the current epoch will promptly stop, preventing any forthcoming rewards and substantially reducing the overall reward count for the drone. The drone will acquire the capacity to proactively evade such incidents by imposing a significant penalty for collisions, as outlined in Eq (8).

$$r_{collision} = \begin{cases} -10 & if\ collision\ occurs \\ 0 & otherwise \end{cases} \quad \text{Eq (8)}$$

The fifth element of the reward formulation is another penalty term for the drone navigating through areas characterized by robust wind patterns, as depicted in Eq (9). This entails administering an adverse reward for each incremental drone movement in a zone of strong winds. This framework is designed to motivate the drone to curtail its presence in these regions, prompting it to seek alternative pathways actively. The structure of this function is linear; hence, the drone's proximity to a wind-prone area correlates with the magnitude of the associated penalty. This mechanism serves as an impetus for the drone to disengage from areas with severe wind conditions.

$$r_{stuck} = \begin{cases} -0.1 \times \frac{dis_{wind}}{20}\ per\ step & if\ in\ strong\ wind\ zone \\ 0 & otherwise \end{cases} \quad \text{Eq (9)}$$

Where $r_{stuck}$ represents the penalty that will be imposed upon the drone, $dis_{wind}$ signifies the spatial gap between the drone's current location and the recorded position of the strong wind zone.

The sixth reward factor is also a penalty when the drone expends its allocated time yet fails to reach the designated target, as illustrated in Eq (10). Battery power consumption is paramount for drones, given their dependence on internal energy sources to fuel their propulsion mechanisms, sensors, cameras, and additional hardware. Suppose the drone consumes a substantial amount of energy during a single flight. In that case, it risks depleting its power reserves and consequently being rendered incapable of fulfilling its objectives or returning to its initial starting point. Given the drone's inability to successfully complete its tasks under such circumstances, a negative penalty is administered to reflect this outcome.

$$r_{timeout} = \begin{cases} -10 & if\ step\ reaches\ maxmium \\ 0 & otherwise \end{cases} \quad \text{Eq (10)}$$

Lastly, the concluding element of the reward scheme is the time-based penalty function applicable to the drone's performance, as delineated in Eq (11). This penalty escalates proportionally with the duration the drone allocates to the mission. The primary objective of this penalty function is to incentivize the drone to opt for the most expeditious route to the target, thereby encouraging the minimization of mission time.

$$r_{time} = -0.001 \; per \; step \qquad \text{Eq (11)}$$

### 3.6. 3D Bounding Box Algorithm

We employ 3D bounding boxes to streamline the storage of obstacle geometry, specifically the outlines of buildings. A 3D bounding box is a rectangular prism encompassing a three-dimensional object, such as a building. This box is characterized by its triad of dimensions - length, width, and height - as well as its spatial positioning and orientation. In computer vision and robotics, the adoption of 3D bounding boxes is pervasive, encapsulating the spatial boundaries of entities within a scene. Broadly speaking, when a 3D box is projected onto an image, it can invariably be enveloped by a 2D box. In this context, Mousavian, et al. [61] introduced an ingenious deep-learning model that forecasts the positioning of 3D bounding boxes by leveraging the outcomes of their 2D counterparts. Their approach establishes a web of relationships between the 2D and 3D bounding box representations, denominated as point-to-side correspondence constraint equations. Exemplifying this, one of these equations is outlined below:

$$x_{min} = \left( K[R \; T] \begin{pmatrix} \frac{d_x}{2} \\ -\frac{d_y}{2} \\ \frac{d_z}{2} \\ 1 \end{pmatrix} \right) \qquad \text{Eq (12)}$$

Where $x_{min}$ stands as a representation of the minimum x-position of the 2D bounding box, $K$ denotes the intrinsic matrix of the camera, $R$ embodies the rotation of the frame, $T$ encapsulates the translation of the structure, $d_x$ corresponds to the length of the 3D bounding box, $d_y$ signifies the width of the 3D bounding box, $d_z$ embodies the height of the 3D bounding box.

Given that a 2D bounding box encompasses four vertex points, which are $x_{min}$, $x_{max}$, $y_{min}$, $y_{max}$, it becomes evident that a quartet of equations mirroring the above vertex can be articulated. These equations collectively form the bedrock of constraints linking the 2D bounding box to its 3D counterpart. This ensemble of equations serves as a foundational framework, and it's important to note that a total of nine parameters necessitate determination within this configuration. Among these parameters, six pertain to the realm of affine transformations, while the remaining three are tethered to the dimensions of the bounding box. Drawing from this conceptual foundation, our endeavor unfolds as we develop the 3D bounding box algorithm grounded in this model. The intricate specifications of this algorithm are illuminated in **Fig. 5**.

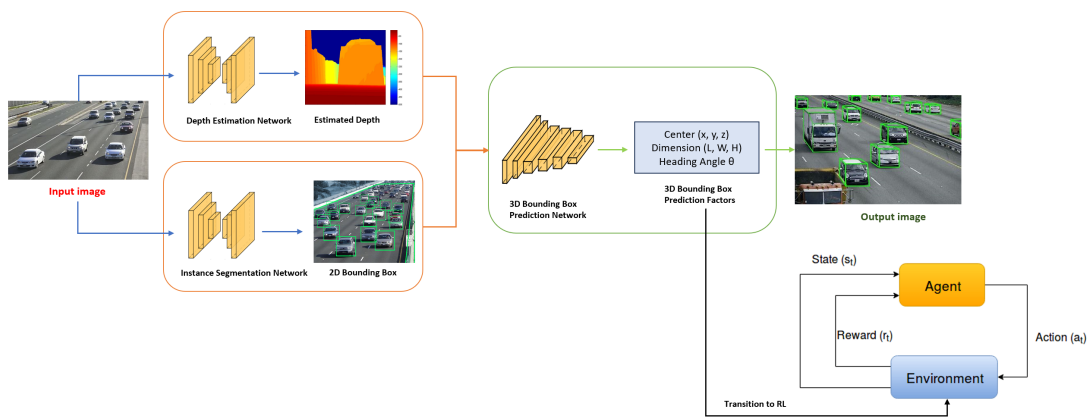

**Fig. 5** The bounding box algorithm used in this study

The drone's onboard camera provides a visual depiction of the obstacles lying ahead. Our design deliberately restricts the volume of obstacles incorporated into the model. This strategic limitation mirrors the genuine constraints that most drones face for data storage and computational capabilities. This controlled scenario enables us to scrutinize the efficacy of the MORL in the face of sparse data inputs. Specifically, when the sensor captures more than three obstacles, the drone will exclusively engage with the three nearest ones. Upon the drone's detection of any obstacle, the initial step involves the application of the 2D bounding box algorithm, an essential conduit to condense the raw data into a more manageable format. This entails the deployment of a 2D detector to identify and encase the obstacles within bounding boxes. Subsequently, these bounding box outcomes are harnessed as inputs for the 3D bounding box network. This neural network is skillfully calibrated to convert the convolutional features into pivotal parameters: the angles and dimensions that govern the obstacles. These parameters feature prominently within the constellation of the four constraint equations.

In the final stage of the pipeline, leveraging the pre-established knowledge of six parameters, we uncover the remaining trio of parameters from the quartet of constraint equations. These parameters, detailing the relative positioning of the 3D bounding box, are deduced with relative ease. Armed with the obstacles' dimensions and relative positions, we calculate all eight vertex coordinates for each obstacle. This ensemble of data is then transferred into the reinforcement learning model.

## 4. EXPERIMENT

In this work, we created our own drone simulator based on previous work to test the feasibility of the proposed reinforcement learning training network. Simulated environments offer a safe and cost-efficient means of training drones. Training algorithms and parameters can be promptly adjusted and refined, while the drone's performance can be assessed and analyzed in real time. The concerns related to real-world experiments such collisions and other safety hazards can be avoided. This expedites learning, resulting in more proficient and capable drones in a shorter timeframe. As a result, we built our drone simulator in Unity 3D due to its capacity to construct a meticulously scaled simulation environment replete with analytical functions for interpreting and generating external experiment data. For a demo, please refer to https://youtu.be/vS9vtHbjhDg.

### 4.1. Drone Simulation

In order to accurately recreate the authentic behavior of a drone in our simulation, it is essential to ensure that all parameters utilized in the drone simulation closely align with their real-world counterparts. For this simulation, we have chosen the DJI Mavic 2 Pro drone, a popular commercial model in the unmanned aerial vehicles (UAVs) market, and it is equipped with camera stabilization systems [62]. This drone features advanced GPS technology, guaranteeing precise positioning and navigation capabilities. These crucial aspects enhance flight stability and allow the possibility of creating intelligent flight modes that simplify the capture of complex aerial shots. Furthermore, we have incorporated a depth camera sensor to enhance data acquisition, replacing the default obstacle avoidance system with the RGB camera. Referring to the official specifications, the drone can achieve a top speed of 45 mph and reach a maximum flight altitude of approximately 1,640 feet. However, considering factors such as urban environments and airspace regulations, we have added restrictions on the drone's height and speed. Taking all these factors into consideration, the drone's specifications for this simulation have been configured as follows:

- Weight: 2 lbs
- Maximum speed: 30 mph
- Maximum altitude: 200 feet

## 4.2. Simulated Environments

The simulated environment represented a district measuring 3,000 × 3,000 feet within an urban setting. We placed 49 tall buildings randomly throughout this simulation, with varying widths and lengths ranging from 130 to 260 feet and heights spanning from 200 to 400 feet. We imposed no physical boundaries on the simulated area, allowing wind to flow freely. In each training epoch, we randomly spawned the drone on one side of the site while the target point was randomly positioned on the opposite side. The drone's mission was to traverse the entire site safely to reach the target. Our simulation granted the drone 3D navigation capabilities, enabling it to adjust its altitude. However, to encourage the drone to navigate through the buildings and dynamic wind zones rather than flying over them, we set a height limit of 200 feet. This constraint compelled the drone to learn strategies for navigating the challenging terrain, and as a result, the primary scenario's dimensions were defined as a 3,000 × 3,000 × 200 feet cube.

As for the wind simulation, we initialized the wind speed at 20 mph. During training, we modulated the wind speed within the 3 to 45 mph range, encompassing common real-life scenarios. Another critical parameter in the wind simulation was the initial wind direction. Since we concentrated on the drone's navigation within the X-Y plane, we created a random horizontal initial wind for each epoch, its direction ranging from 0 to 360 degrees to the whole simulation environment, as illustrated in **Fig. 6**. In each training epoch, the drone agent encountered one particular wind direction and wind speed randomly, introducing variability into the simulated environment. Depending on the drone's flight direction and the initial wind direction, the wind's impact on the drone differed. For example, when the part of wind direction paralleled the drone's flight path, it would increase the drone's speed. Managing a higher speed posed a challenge for automatic navigation control, as it allowed for only a brief collision avoidance window. Another challenging scenario arose with crosswinds, where the wind direction was perpendicular to the drone's flight direction. In such cases, maintaining the desired path or implementing stabilization controls presented nontrivial challenges for the control algorithm. The goal of the proposed MORL was to teach the drone to discern the underlying patterns in wind effects and spontaneously generate optimal solutions to counteract its influence.

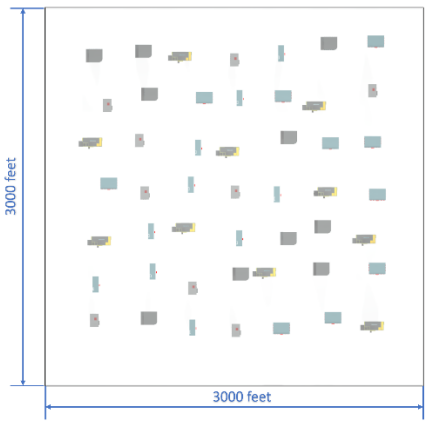
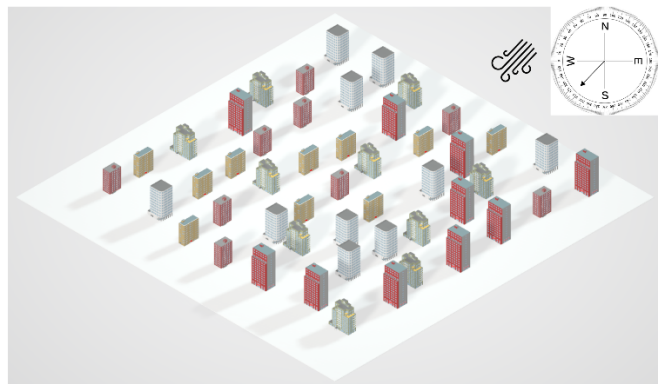

(a) The layout of the scene  (b) The simulation of wind directions

**Fig. 6** Example of the simulated environments and drone

## 4.3. Training Process

We implemented a curriculum learning approach to facilitate the drone's acquisition of complex skills, including navigating challenging wind zones. Curriculum learning is a well-established and effective machine learning technique designed to train models by gradually increasing the complexity of their tasks [63]. The fundamental concept behind curriculum learning is to expose the model to tasks, commencing with relatively straightforward ones and progressively introducing more challenging objectives over time. By adhering to this structured curriculum, the model can learn more efficiently and effectively while avoiding being trapped in suboptimal solutions. Various metrics, such as loss, entropy, or mean final reward, can be employed to determine the appropriate number of epochs required for each training phase. This approach draws inspiration from human learning, where individuals begin with simple concepts and incrementally build upon them to tackle more intricate concepts. In our study, the drone has multiple objectives, including navigation, obstacle avoidance, wind zone control, and detouring around strong wind zones. It is difficult for the drone to learn these aspects in a single training. Therefore, we employed a curriculum learning strategy and divided the training process into four phases.

In phase 1, the primary aim was to train the drone in navigation and obstacle avoidance. We only introduced building obstacles into the training environment without wind simulation. Drone and target positions and the distribution of obstacles were randomized in each epoch. The intent was to prevent the drone from memorizing obstacle locations. During training, collision incidents or the drone flying out of the scenario led to negative rewards and epoch termination. A timeout penalty was imposed if the step limit was reached, set at 50,000 steps. We advanced to the next phase once a high success rate was achieved.

During phase 2, we connected the machine learning model and the reinforcement learning training environment. Since the drone agent had already built up its navigation and obstacle avoidance system, we introduced a wind zone into the environment in this phase. The primary objective was to teach the agent to control the drone in windy conditions. Aside from the wind simulation, all other settings remained consistent with Phase 1. In each epoch, a random wind direction will be generated in a Python program and then used as the initial condition for the CNN model, which was used to calculate the CFD approximation result for the wind zone around each building.

While the drone could now navigate safely in an environment with random wind zones, it could still become stuck in certain situations due to the complexity of wind zones. To address this, we introduced more dynamic training conditions in Phase 3. The wind direction choices ensured that all initial wind directions were opposite to the drone's flight direction. Furthermore, we employed a new training strategy in which the environmental factors and the drone's status were retained between epochs. Combined with a specific penalty function, this strategy aimed to help the drone escape from local traps and find solutions for infinite loop situations.

The final phase integrated all the features from the previous training phases to enhance the model's overall performance. One of the advantages of curriculum learning is that it preserves previously learned knowledge. However, it is crucial to strike a balance, as overemphasizing a single sub-training criterion can lead to overfitting for a specific task and render the model unsuitable for multi-objective scenarios. During this phase, the model was retrained to achieve superior performance in a comprehensive environment. Additionally, a variable learning rate was employed to mitigate issues related to local optima and saddle points, facilitating escape from such traps.

### 4.4. Computational Fluid Dynamics Setting

Our research utilized Computational Fluid Dynamics (CFD) data to assess our algorithm's performance rigorously. We employed the OpenFOAM software package to provide a concise overview of the CFD simulations. This allowed us to define the simulation domain, boundary conditions, and relevant physical parameters illustrated in **Fig. 7**.

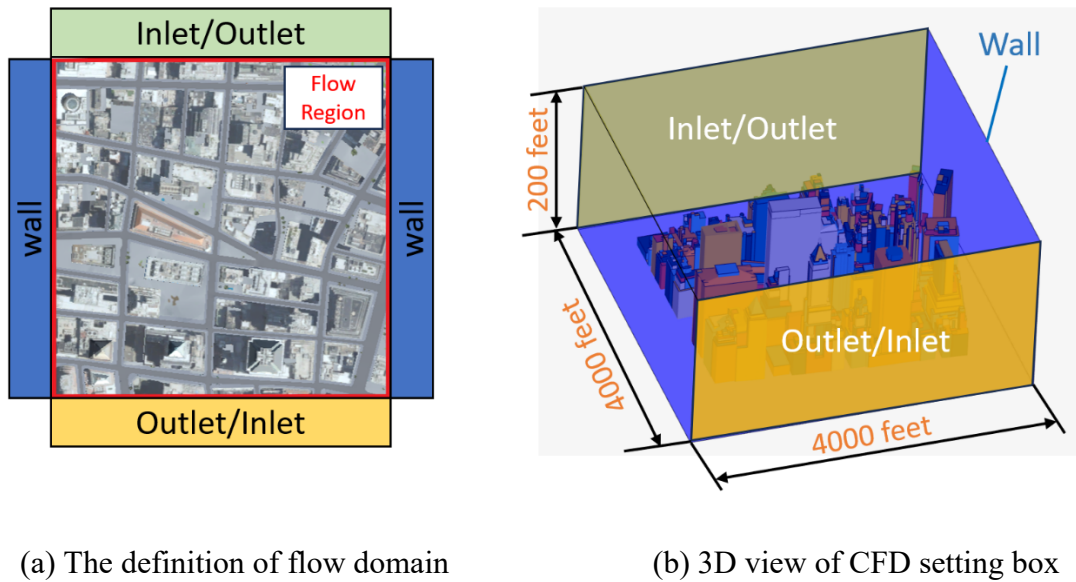

(a) The definition of flow domain  (b) 3D view of CFD setting box

**Fig. 7** Details of CFD setting

In **Fig. 7(a)**, we delineate the fluid domain and outline the boundary conditions within the CFD environment. The fluid in the computational field is represented as air, with a designated density of 1.196 kg/m³. We consider four distinct wind conditions in our validation scenarios, each with two different wind directions. For each wind direction, we specify inlet wind speeds of 5 m/s and 10 m/s, while the air pressure at the outlet is set at 0.015 kPa and 0.0625 kPa, respectively.

## 5. RESULTS

### 5.1. Simulated Cases

We conducted a series of tests to assess the effectiveness of the MORL method in guiding a drone to locate the final target while navigating and evading dynamic wind zones. **Fig. 8** illustrated the drone agent's learning process, which involved learning navigation and collision avoidance in the training environment during Phase 1. In this visualization, the target was represented by the blue sphere, the red rectangle corresponded to the bounding box result, and the buildings were cubes. To depict the drone's path, we employed a purple line to trace its trajectory in space. **Fig. 8(a)** showcased the drone's behavior during the initial stages of training. At this point, the agent lacked an understanding of its task and the significance of the inputs, leading it to explore the environment randomly and frequently collide with obstacles and structures. In **Fig. 8(b)**, we observed that the drone had acquired knowledge of its task, commencing its movement towards the target based on past rewards. It exhibited controlled flight within the training environment's boundaries and avoided collisions with structures. However, it still hit the buildings due to its incomplete comprehension of the input data, which represented the locations and dimensions of the building obstacles. In **Fig. 8(c)**, we found that though the drone successfully flew across the building area, it still did not know the target of its flight. It kept hovering around the target. **Fig. 8(d)** demonstrated that the drone successfully reached the target while circumventing all obstacles, achieving a remarkable 95% success rate after 80,000 epochs. Furthermore, the drone had learned to select the shortest path based on local information, efficiently navigating from the starting point to the target.

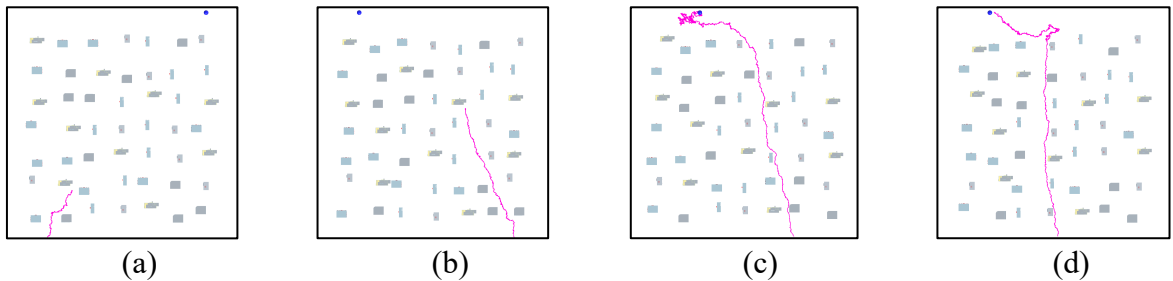

(a)            (b)            (c)            (d)

**Fig. 8** Examples of training results in Phase 1

**Fig. 9(a)** presented an example of the drone's poor performance during the early stages of training in a scenario featuring wind zones. Unlike the previously high success rate observed in a windless environment, the drone faced significant challenges during the initial training phase. At this point, it continued to employ strategies that had worked previously when wind was not a factor. However, the wind forced the drone off its intended course, ultimately leading to collisions with buildings. **Fig. 9(b)** illustrated the drone's evolving recognition of the impact of wind, prompting it to adjust its decision-making process. In **Fig. 9(c)**, we observed that the drone could now safely reach the target while navigating around obstacles and through wind zones. This achievement corresponded to a 90% success rate after 80,000 training epochs. However, it's worth noting that approximately 10% of the total cases resulted in timeouts during training Phase 2. **Fig. 9(d)** offered insight into the drone's behavior during timeout instances. As depicted, the drone became trapped between two narrow buildings, indicating that the wind's magnitude in this region exceeded the drone's maximum speed. In essence, if the drone attempted to enter this area with wind blowing against its intended direction of movement, it got pushed back. Upon returning to its previous position, the drone still decided to move forward due to consistent environmental information, resulting in an endless loop of local entrapment.

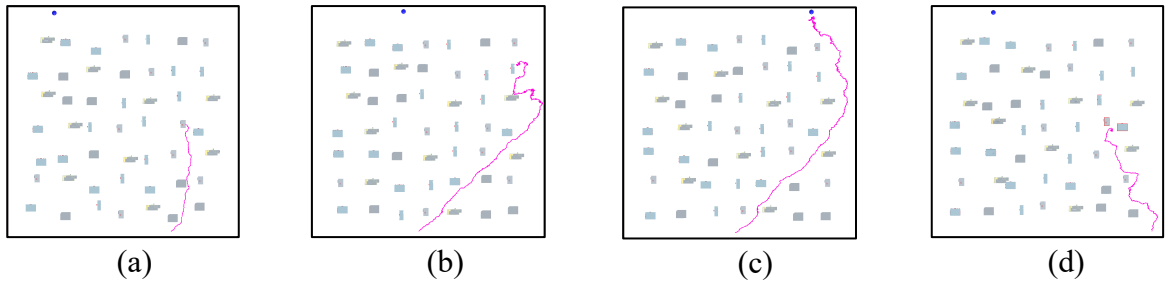

(a)　　　　　　　(b)　　　　　　　(c)　　　　　　　(d)

**Fig. 9** Examples of training results in Phase 2

**Fig. 10** depicted the progressive learning process of the drone agent as it acquired the ability to detour around and evade a formidable strong wind zone. During training Phase 3, we implemented a retraining strategy to increase the likelihood of timeout scenarios deliberately. In this approach, the drone was retrained in the same environment whenever a timeout occurred. Due to the capabilities of the LSTM network, the drone could effectively recall its prior actions, enabling it to navigate around the strong wind area successfully. **Fig. 10(a)** illustrated the drone's initial attempts to identify an alternative route away from the strong wind zone, though it faced challenges during the initial stages of training. A substantial penalty associated with lingering near the strong wind zone compelled the drone to explore different flight directions in an attempt to detour. However, the drone would sometimes return to the vicinity of the penalty area when moving beyond its boundaries. **Fig. 10(b)** demonstrated that with sufficient training time spent near the strong wind zone, the drone eventually discovered an alternate path. This milestone signified the drone's ability to recognize the presence of a strong wind zone, even without the assistance of an aerodynamic force sensor, and adapt its decision-making based on its past actions.

**Fig. 10(c)** and **Fig. 10(d)** showcased the drone's progress as it learned to swiftly react upon entering a strong wind zone, opting to change its course to navigate around it effectively. Following an adequate number of training epochs, the drone displayed the capability to make rapid adjustments when encountering such zones and actively chose to bypass them. In addition, the drone acquired the ability to fly close to the buildings, a behavior that may have appeared unconventional by traditional standards. Based on the CFD results, this adaptation could be attributed to the drone's comprehension of wind patterns. In these simulations, it became evident that the wind force diminished significantly when the location was near the buildings, influencing the drone's behavior. In this training environment, we achieved an impressive 98% success rate after 100,000 epochs.

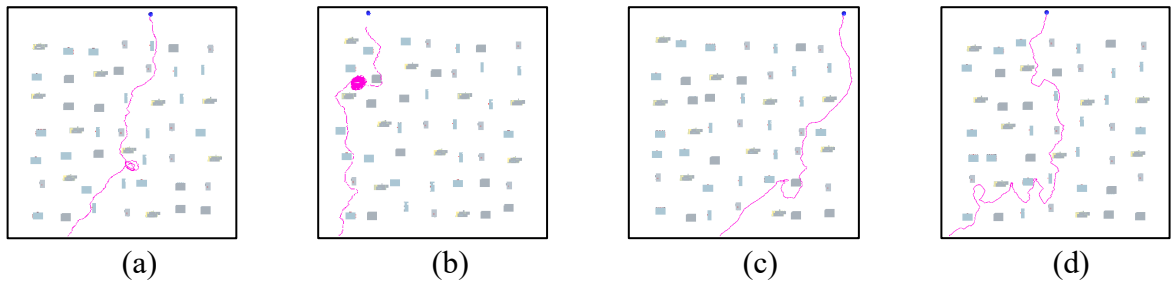

(a)　　　　　　(b)　　　　　　(c)　　　　　　(d)

**Fig. 10** Examples of training results in Phase 3 and 4

**5.2. Comparison between the different CFD data methodologies**

**Fig. 11** displayed the outcomes of various wind simulation techniques applied to a single building. In **Fig. 11(a)**, we observed the results of mathematical aerodynamic representation modeling, a method employed in our prior research. This approach offered the advantage of being computationally efficient, imposing minimal additional workload when integrated with reinforcement learning. However, it could only be applied when the wind was parallel or perpendicular to the obstacles' directions. Besides, its performance fell short of the CFD approximation results obtained through Convolutional Autoencoder, as illustrated in **Fig. 11(b)**. Convolutional Autoencoder enabled the acquisition of high-accuracy CFD data in less time than traditional Navier–Stokes equation solvers. This method's data transfer delay between software components was one drawback. To address this challenge, we introduced MessagePack, a binary serialization format designed for enhanced compactness and efficiency compared to JSON for data encoding during transmission and storage. In **Fig. 11(c)**, we presented the ground truth data generated by OpenFOAM, representing the most accurate means of obtaining high-precision wind flow data. However, this method came at the cost of time, necessitating approximately 30 minutes to generate results for a single environmental case.

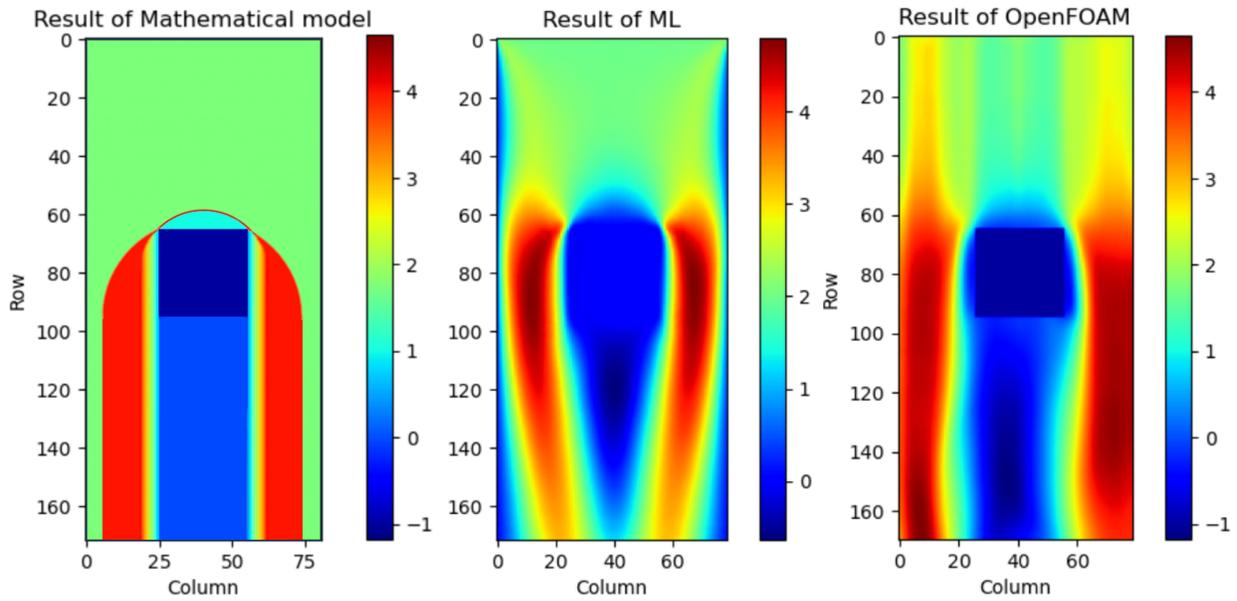

(a) (b) (c)

**Fig. 11** The results of different wind simulation methods

**Table 1** lists the difference between each methodology.

|  | Mathematical modeling | Convolutional Autoencoder | OpenFOAM |
|---|---|---|---|
| Time cost for one case | Less than 1 second | 5 seconds | 30 minutes |
| Data transfer delay | No | 1 seconds | 5 seconds |
| Data precision | low | medium | high |
| Applicability | low | high | high |
| Effected scope | small | large | large |

**Table.1** The comparison of results between different methodologies

**Fig. 12** showed the data collection from each training epoch, recording the mean reward every 100,000 steps. In **Fig. 12(a)**, we presented the learning curves using various wind simulation methodologies. The graph illustrated that using a convolutional autoencoder model resulted in a faster learning rate than the mathematical model. Additionally, the agent demonstrated superior performance in cumulative rewards, indicating that the drone agent could effectively navigate through challenging wind conditions during its journey. **Fig. 12(b)** displayed the cumulative rewards curve during training phases 3 and 4. While the drone's behavior trained with the mathematical wind simulation methodology initially lagged behind that of the convolutional autoencoder model, it ultimately achieved higher cumulative rewards. This outcome could be attributed to the inherent complexity differences between these two models. Although the mathematical model offered computational efficiency and required fewer resources, its results provided less detailed information compared to the convolutional model, potentially limiting the applicability of the autonomous drone model.

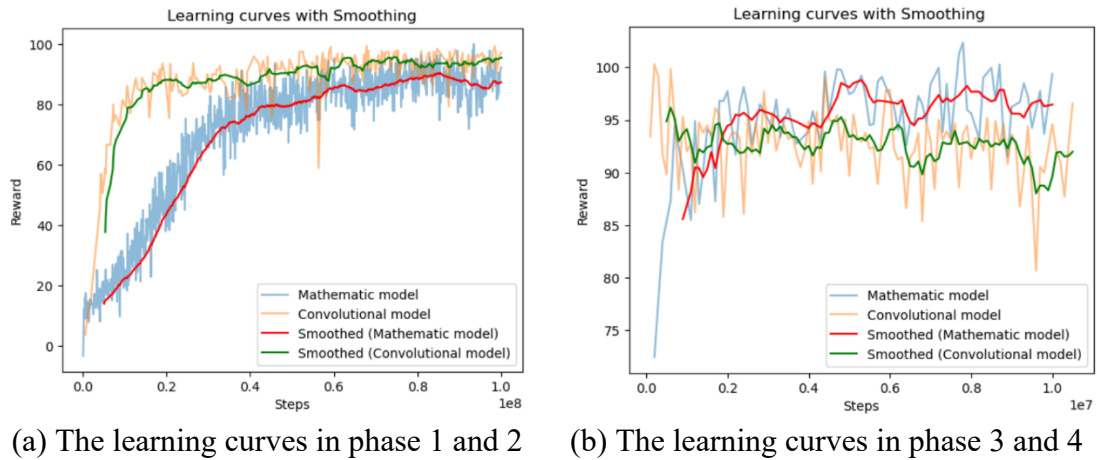

(a) The learning curves in phase 1 and 2    (b) The learning curves in phase 3 and 4

**Fig. 12** The learning curve of all the phases

## 5.3. Validation Study

To validate the effectiveness of the proposed MORL method in guiding a drone through a more authentic urban setting enriched with precise aerodynamic data, we utilized the Manhattan district within New York City as our validation case, as depicted in **Fig. 13**. In Manhattan, the prevailing wind direction typically originated from the northwest, primarily due to the city's location on the east coast of the United States and its exposure to prevailing westerly winds. Consequently, we had deliberately selected two opposing wind directions, namely northwest and southeast, as the primary wind directions for assessing the algorithm's performance. **Fig. 13 (c)** and **(d)** offered a first-person perspective from the drone's vantage point within the virtual simulator.

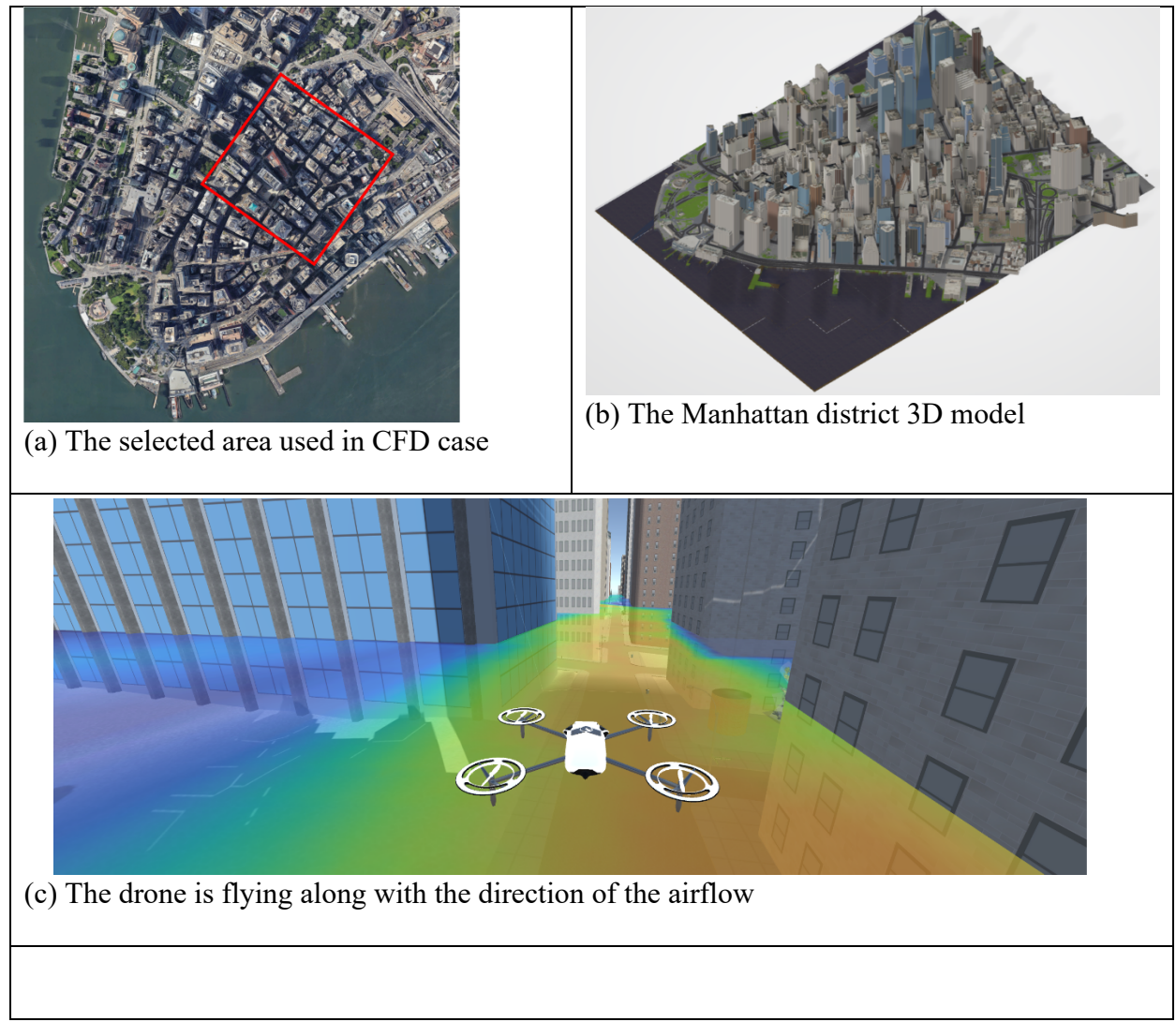

(a) The selected area used in CFD case

(b) The Manhattan district 3D model

(c) The drone is flying along with the direction of the airflow

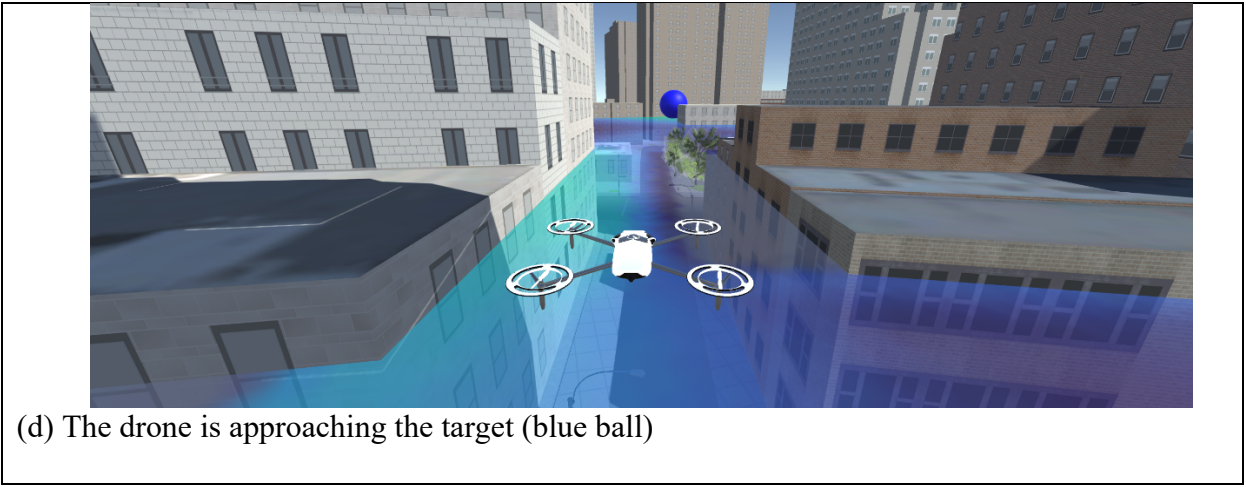
(d) The drone is approaching the target (blue ball)

**Fig. 13 The details of the Manhattan district model**

There are two primary distinctions between the training environment and the validation test case. The first distinction lies in the environmental layout. The Unity program randomly generates and distributes building layouts in the training environment. In contrast, the validation case employs a real-city model, the Manhattan district in New York City, to assess the drone's behavior. The second difference pertains to the method of wind simulation. During training, we employ a Convolutional Autoencoder to simulate the wind map. However, we utilize OpenFOAM to generate CFD results for the Manhattan model in the validation case, which offers higher authenticity in wind flow data.

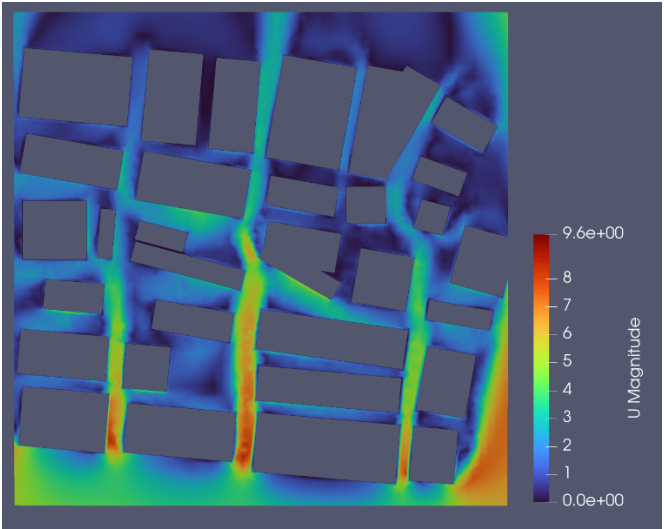

(a) The CFD case 1 (wind from SE to NW, $v_{wind}$= 5m/s)

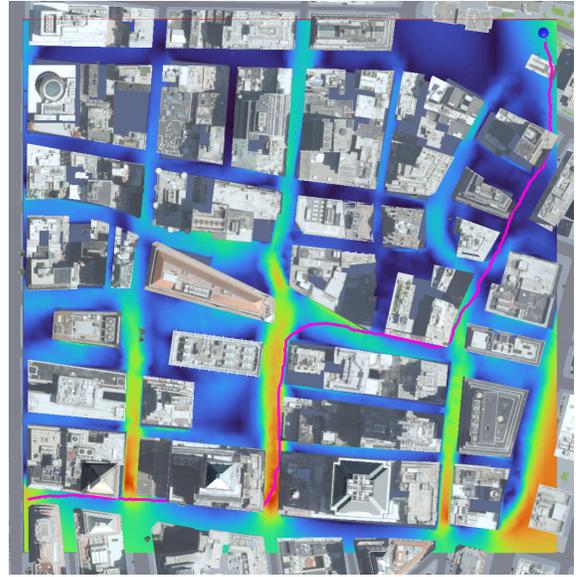

(b) The route of drone in case 1

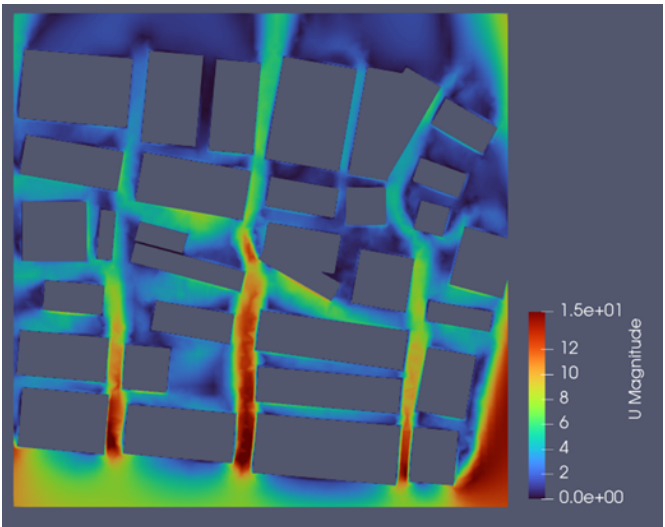

(c) The CFD case 2 (wind from SE to NW, $v_{wind}$= 10m/s)

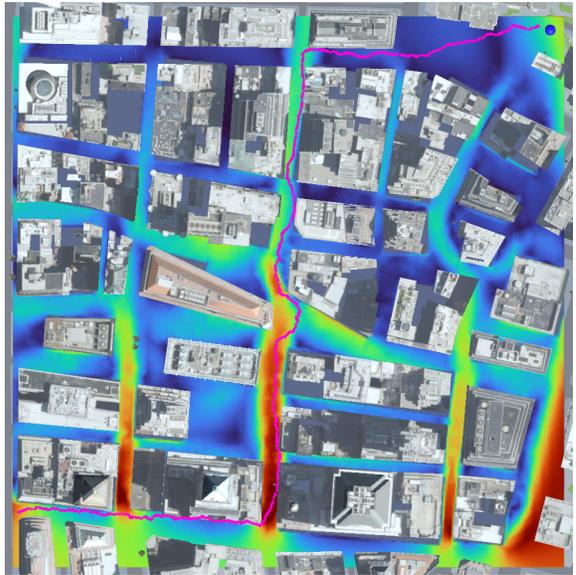

(d) The route of drone in case 2

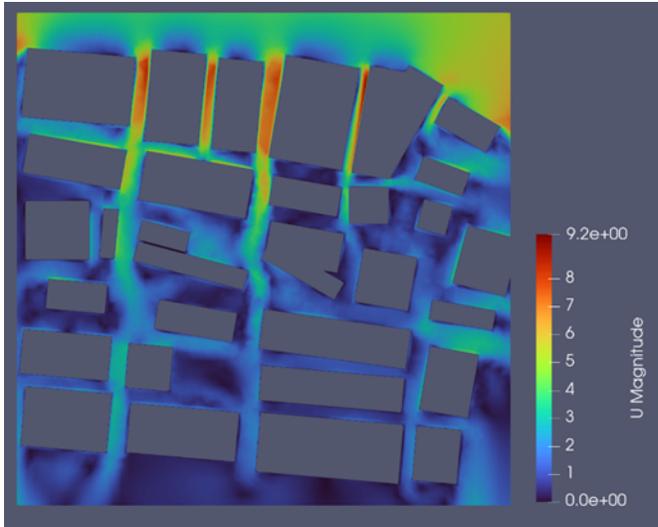

(e) The CFD case 3 (wind from SE to NW, $v_{wind}$= 5m/s)

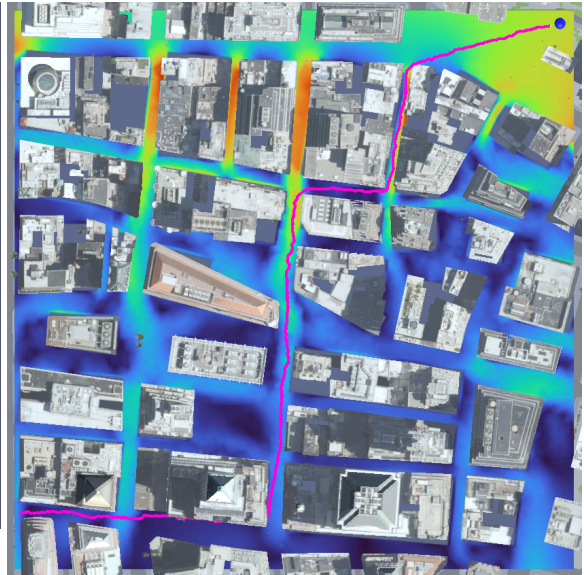

(f) The route of drone in case 3

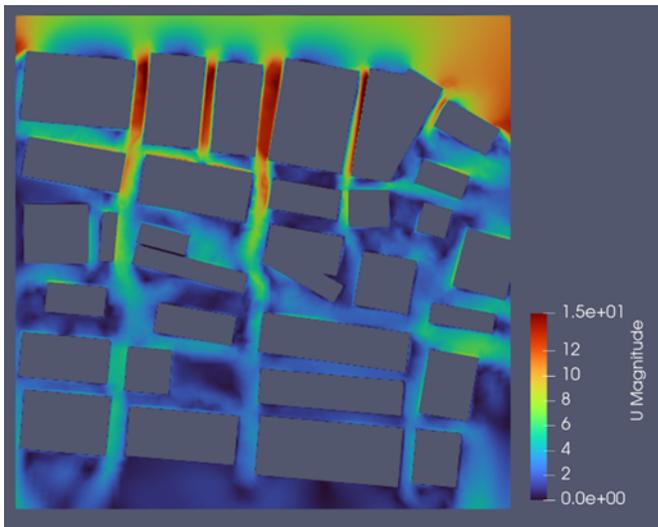

(g) The CFD case 4 (wind from SE to NW, $v_{wind}$= 10m/s)

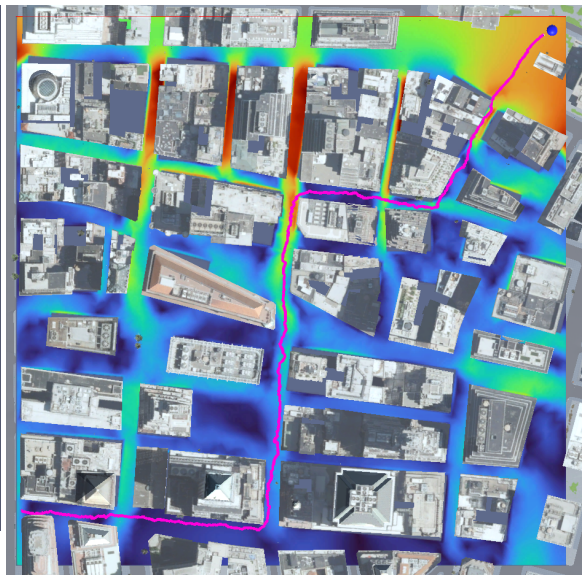

(h) The route of drone in case 4

**Fig. 14 The validation case study results**

**Fig. 14** illustrates the drone's adaptive behaviors within the Manhattan model, where it grapples with two competing objectives: selecting the most efficient route to the target while navigating around areas of high wind intensity. These results effectively showcase the ability of our proposed algorithm to train the drone to adapt to wind dynamics and adeptly navigate through unfamiliar terrain. **Fig. 14** highlights the drone's behavior under calm wind conditions.

The drone will always choose the shortest path between its starting point and destination in windless conditions. However, **Fig. 14(b)** and **(d)** depict differing route selections made by the drone in response to varying wind strengths. Specifically, as the wind direction shifts from southeast to northwest, favorably assisting the drone's flight, the preference for a longer yet more direct route emerges, as evident in **Fig. 14(d)**. On the contrary, **Fig. 14(f)** and **(h)** showcase alternative route choices when the wind forces counter the drone's movement. In these scenarios, the presence of potent wind zones, indicated by the red segments in **Fig. 14(e)** and **(g)**, obstructs the drone's ability to follow the shortest path observed in **Fig. 14(b)**. To mitigate the impact of adverse wind forces, the drone strategically selects routes that pass through milder wind zones, relying solely on imagery data and previously memorized positions from earlier epochs. This deviation from conventional paths underscores the drone's calculated trade-off decision, opting to cover more distance to minimize the detrimental influence of the environmental variable—wind force.

## 6. CONCLUSIONS

In this study, we introduce an advanced approach for drone navigation in urban settings, considering the influence of wind. Our method relies on multi-objective DRL, LSTM networks, and a Convolutional Autoencoder. We emphasize that traditional drone navigation and obstacle avoidance systems often overlook the impact of environmental factors, particularly the lateral wind patterns prevalent around tall buildings in urban areas. Furthermore, our research harnesses the power of machine learning to generate wind data for training, a strategy that significantly reduces computational and time costs compared to traditional CFD simulations. Despite the computational efficiency, our machine learning-generated wind data also maintains a high level of quality. Besides, we have developed an innovative data transfer structure between RL and ML for simulating wind distribution around buildings to improve training efficiency and create a training environment based on this methodology. Specifically, we utilized an LSTM architecture with a Proximal Policy Optimization (PPO) policy to train the drone to navigate and evade building obstacles in a wind-free environment. Subsequently, we introduced simulated wind zones into the environment, instructing the drone to perform detours effectively. Lastly, we validated the drone's navigation and obstacle avoidance capabilities within a setting featuring varying wind conditions. This validation involved integrating the model with an actual New York City urban model complemented by Computational Fluid Dynamics (CFD) results. Our findings affirm that our proposed system adeptly adapts to wind effects in unfamiliar environments, reaching its destination safely without collisions. Furthermore, the drone's flight path illustrates the agent's ability to plan an optimal global route using local information, eliminating the necessity for a global map. Additionally, we have demonstrated the feasibility of transferring a drone model trained with simulation wind methodology to an unknown environment supported by CFD results. The agent effectively learns to control the drone under wind effects from the wind zones created by the simulation methodology.

While our proposed system shows promising results, several limitations require further investigation. The current approach primarily focuses on wind as the dominant environmental factor affecting drone navigation. Other factors like precipitation, temperature, signal strength, and air pressure, which could significantly impact drone performance, are not currently accounted for and warrant consideration. Secondly, the simulation environment assumes a static urban setting and does not factor in dynamic elements like moving vehicles or pedestrians. Incorporating these dynamic elements could provide a more accurate representation of real-world conditions and enhance the robustness of our system. Thirdly, the reinforcement learning model we employ may face challenges when applied to larger, more complex environments or diverse drone models. In addition, the current Convolutional Autoencoder is good enough to generate the CFD approximation result for a single building. However, the simple linear combination of those single buildings differs from the ground truth result when it comes to the multiple buildings scenario. Investigating the scalability of our Convolutional Autoencoder system to accommodate various and more complex urban settings is a crucial topic for future research. To address these limitations, future research should focus on optimizing the wind simulation methodology, incorporating additional environmental factors, and accounting for dynamic elements within urban environments. Furthermore, improving the scalability of our proposed system and optimizing the training process to reduce computational demands will be essential in enhancing the practical applicability of this approach.

## ACKNOWLEDGEMENTS

This material is supported by the Air Force Office of Scientific Research (AFOSR) under grant FA9550-22-1-0492. Any opinions, findings, conclusions, or recommendations expressed in this article are those of the authors and do not reflect the views of the AFOSR.

**References:**

[1] A. Restas, "Drone applications for supporting disaster management," *World Journal of Engineering and Technology,* vol. 3, no. 03, p. 316, 2015.

[2] B. Mishra, D. Garg, P. Narang, and V. Mishra, "Drone-surveillance for search and rescue in natural disaster," *Computer Communications,* vol. 156, pp. 1-10, 2020.

[3] A. Capolupo, S. Pindozzi, C. Okello, N. Fiorentino, and L. Boccia, "Photogrammetry for environmental monitoring: The use of drones and hydrological models for detection of soil contaminated by copper," *Science of the Total Environment,* vol. 514, pp. 298-306, 2015.

[4] M. Esposito, M. Crimaldi, V. Cirillo, F. Sarghini, and A. Maggio, "Drone and sensor technology for sustainable weed management: A review," *Chemical and Biological Technologies in Agriculture,* vol. 8, no. 1, pp. 1-11, 2021.

[5] A. Entrop and A. Vasenev, "Infrared drones in the construction industry: designing a protocol for building thermography procedures," *Energy procedia,* vol. 132, pp. 63-68, 2017.

[6] D. Ventura, M. Bruno, G. J. Lasinio, A. Belluscio, and G. Ardizzone, "A low-cost drone based application for identifying and mapping of coastal fish nursery grounds," *Estuarine, Coastal and Shelf Science,* vol. 171, pp. 85-98, 2016.

[7] D. Floreano and R. J. Wood, "Science, technology and the future of small autonomous drones," *nature,* vol. 521, no. 7553, pp. 460-466, 2015.

[8] S. Ragi and E. K. Chong, "UAV path planning in a dynamic environment via partially observable Markov decision process," *IEEE Transactions on Aerospace and Electronic Systems,* vol. 49, no. 4, pp. 2397-2412, 2013.

[9] S. M. Shavarani, M. G. Nejad, F. Rismanchian, and G. Izbirak, "Application of hierarchical facility location problem for optimization of a drone delivery system: a case study of Amazon prime air in the city of San Francisco," *The International Journal of Advanced Manufacturing Technology,* vol. 95, pp. 3141-3153, 2018.

[10] M. T. R. Khan, M. Muhammad Saad, Y. Ru, J. Seo, and D. Kim, "Aspects of unmanned aerial vehicles path planning: Overview and applications," *International Journal of Communication Systems,* vol. 34, no. 10, p. e4827, 2021.

[11] Y. Liu, Q. Wang, H. Hu, and Y. He, "A novel real-time moving target tracking and path planning system for a quadrotor UAV in unknown unstructured outdoor scenes," *IEEE Transactions on Systems, Man, and Cybernetics: Systems,* vol. 49, no. 11, pp. 2362-2372, 2018.

[12] B. Blocken, T. Stathopoulos, and J. Carmeliet, "Wind environmental conditions in passages between two long narrow perpendicular buildings," *Journal of Aerospace Engineering,* vol. 21, no. 4, pp. 280-287, 2008.

[13] L. Biao, J. Cunyan, W. Lu, C. Weihua, and L. Jing, "A parametric study of the effect of building layout on wind flow over an urban area," *Building and Environment,* vol. 160, p. 106160, 2019.

[14] G. Calzolari and W. Liu, "Deep learning to replace, improve, or aid CFD analysis in built environment applications: A review," *Building and Environment,* vol. 206, p. 108315, 2021.

[15] V. Puri, A. Nayyar, and L. Raja, "Agriculture drones: A modern breakthrough in precision agriculture," *Journal of Statistics and Management Systems,* vol. 20, no. 4, pp. 507-518, 2017.

[16]   L. Apvrille, T. Tanzi, and J.-L. Dugelay, "Autonomous drones for assisting rescue services within the context of natural disasters," in *2014 XXXIth URSI General Assembly and Scientific Symposium (URSI GASS)*, 2014: IEEE, pp. 1-4.

[17]   A. Devos, E. Ebeid, and P. Manoonpong, "Development of autonomous drones for adaptive obstacle avoidance in real world environments," in *2018 21st Euromicro conference on digital system design (DSD)*, 2018: IEEE, pp. 707-710.

[18]   J. Linchant, J. Lisein, J. Semeki, P. Lejeune, and C. Vermeulen, "Are unmanned aircraft systems (UAS s) the future of wildlife monitoring? A review of accomplishments and challenges," *Mammal Review,* vol. 45, no. 4, pp. 239-252, 2015.

[19]   J. Kwak and Y. Sung, "Autonomous UAV flight control for GPS-based navigation," *IEEE Access,* vol. 6, pp. 37947-37955, 2018.

[20]   A. Shahoud, D. Shashev, and S. Shidlovskiy, "Visual navigation and path tracking using street geometry information for image alignment and servoing," *Drones,* vol. 6, no. 5, p. 107, 2022.

[21]   T. Elmokadem and A. V. Savkin, "A hybrid approach for autonomous collision-free UAV navigation in 3D partially unknown dynamic environments," *Drones,* vol. 5, no. 3, p. 57, 2021.

[22]   F. Vanegas, K. J. Gaston, J. Roberts, and F. Gonzalez, "A framework for UAV navigation and exploration in GPS-denied environments," in *2019 ieee aerospace conference*, 2019: IEEE, pp. 1-6.

[23]   M. Y. Arafat, M. M. Alam, and S. Moh, "Vision-Based Navigation Techniques for Unmanned Aerial Vehicles: Review and Challenges," *Drones,* vol. 7, no. 2, p. 89, 2023.

[24]   L. von Stumberg, V. Usenko, J. Engel, J. Stückler, and D. Cremers, "From monocular SLAM to autonomous drone exploration," in *2017 European Conference on Mobile Robots (ECMR)*, 2017: IEEE, pp. 1-8.

[25]   A. Suleiman, Z. Zhang, L. Carlone, S. Karaman, and V. Sze, "Navion: A 2-mw fully integrated real-time visual-inertial odometry accelerator for autonomous navigation of nano drones," *IEEE Journal of Solid-State Circuits,* vol. 54, no. 4, pp. 1106-1119, 2019.

[26]   M. Bolognini and L. Fagiano, "Lidar-based navigation of tethered drone formations in an unknown environment," *IFAC-PapersOnLine,* vol. 53, no. 2, pp. 9426-9431, 2020.

[27]   S. Lee, D. Har, and D. Kum, "Drone-assisted disaster management: Finding victims via infrared camera and lidar sensor fusion," in *2016 3rd Asia-Pacific World Congress on Computer Science and Engineering (APWC on CSE)*, 2016: IEEE, pp. 84-89.

[28]   T. Lee, S. Mckeever, and J. Courtney, "Flying free: A research overview of deep learning in drone navigation autonomy," *Drones,* vol. 5, no. 2, p. 52, 2021.

[29]   M. A. Arshad *et al.*, "Drone Navigation Using Region and Edge Exploitation-Based Deep CNN," *IEEE Access,* vol. 10, pp. 95441-95450, 2022.

[30]   M. Zhang, M. Zhang, Y. Chen, and M. Li, "IMU data processing for inertial aided navigation: A recurrent neural network based approach," in *2021 IEEE International Conference on Robotics and Automation (ICRA)*, 2021: IEEE, pp. 3992-3998.

[31]   N. El-Sheimy and Y. Li, "Indoor navigation: State of the art and future trends," *Satellite Navigation,* vol. 2, no. 1, pp. 1-23, 2021.

[32]	P. Chakravarty, K. Kelchtermans, T. Roussel, S. Wellens, T. Tuytelaars, and L. Van Eycken, "CNN-based single image obstacle avoidance on a quadrotor," in *2017 IEEE international conference on robotics and automation (ICRA)*, 2017: IEEE, pp. 6369-6374.

[33]	M. F. Sani and G. Karimian, "Automatic navigation and landing of an indoor AR. drone quadrotor using ArUco marker and inertial sensors," in *2017 international conference on computer and drone applications (IConDA)*, 2017: IEEE, pp. 102-107.

[34]	L. R. García Carrillo, A. E. Dzul López, R. Lozano, and C. Pégard, "Combining stereo vision and inertial navigation system for a quad-rotor UAV," *Journal of intelligent & robotic systems,* vol. 65, no. 1-4, pp. 373-387, 2012.

[35]	Z. Wang, Y. Wu, and Q. Niu, "Multi-sensor fusion in automated driving: A survey," *Ieee Access,* vol. 8, pp. 2847-2868, 2019.

[36]	Q. Sun, M. Li, T. Wang, and C. Zhao, "UAV path planning based on improved rapidly-exploring random tree," in *2018 Chinese control and decision conference (CCDC)*, 2018: IEEE, pp. 6420-6424.

[37]	N. Wen, X. Su, P. Ma, L. Zhao, and Y. Zhang, "Online UAV path planning in uncertain and hostile environments," *International journal of machine learning and cybernetics,* vol. 8, pp. 469-487, 2017.

[38]	A. Albert, F. S. Leira, and L. S. Imsland, "UAV path planning using MILP with experiments," 2017.

[39]	C. Huang, X. Zhou, X. Ran, J. Wang, H. Chen, and W. Deng, "Adaptive cylinder vector particle swarm optimization with differential evolution for UAV path planning," *Engineering Applications of Artificial Intelligence,* vol. 121, p. 105942, 2023.

[40]	L. P. Kaelbling, M. L. Littman, and A. W. Moore, "Reinforcement learning: A survey," *Journal of artificial intelligence research,* vol. 4, pp. 237-285, 1996.

[41]	V. Mnih *et al.*, "Human-level control through deep reinforcement learning," *nature,* vol. 518, no. 7540, pp. 529-533, 2015.

[42]	H. X. Pham, H. M. La, D. Feil-Seifer, and L. V. Nguyen, "Autonomous uav navigation using reinforcement learning," *arXiv preprint arXiv:1801.05086,* 2018.

[43]	V. J. Hodge, R. Hawkins, and R. Alexander, "Deep reinforcement learning for drone navigation using sensor data," *Neural Computing and Applications,* vol. 33, pp. 2015-2033, 2021.

[44]	G. Muñoz, C. Barrado, E. Çetin, and E. Salami, "Deep reinforcement learning for drone delivery," *Drones,* vol. 3, no. 3, p. 72, 2019.

[45]	A. Ramezani Dooraki and D.-J. Lee, "A multi-objective reinforcement learning based controller for autonomous navigation in challenging environments," *Machines,* vol. 10, no. 7, p. 500, 2022.

[46]	A. Shantia *et al.*, "Two-stage visual navigation by deep neural networks and multi-goal reinforcement learning," *Robotics and Autonomous Systems,* vol. 138, p. 103731, 2021.

[47]	C. Paz, E. Suárez, C. Gil, and J. Vence, "Assessment of the methodology for the CFD simulation of the flight of a quadcopter UAV," *Journal of Wind Engineering and Industrial Aerodynamics,* vol. 218, p. 104776, 2021.

[48] C. Qu, F. B. Sorbelli, R. Singh, P. Calyam, and S. K. Das, "Environmentally-aware and energy-efficient multi-drone coordination and networking for disaster response," *IEEE Transactions on Network and Service Management,* 2023.

[49] S. Giersch, O. El Guernaoui, S. Raasch, M. Sauer, and M. Palomar, "Atmospheric flow simulation strategies to assess turbulent wind conditions for safe drone operations in urban environments," *Journal of Wind Engineering and Industrial Aerodynamics,* vol. 229, p. 105136, 2022.

[50] S. Jeong, K. You, and D. Seok, "Hazardous flight region prediction for a small UAV operated in an urban area using a deep neural network," *Aerospace Science and Technology,* vol. 118, p. 107060, 2021.

[51] T. Chu, M. J. Starek, J. Berryhill, C. Quiroga, and M. Pashaei, "Simulation and characterization of wind impacts on sUAS flight performance for crash scene reconstruction," *Drones,* vol. 5, no. 3, p. 67, 2021.

[52] R. K. Vuppala and K. Kara, "A Novel Approach in Realistic Wind Data Generation for The Safe Operation of Small Unmanned Aerial Systems in Urban Environment," in *AIAA AVIATION 2021 FORUM*, 2021, p. 2505.

[53] J. Milla-Val, C. Montañés, and N. Fueyo, "Economical microscale predictions of wind over complex terrain from mesoscale simulations using machine learning," *Modeling Earth Systems and Environment,* pp. 1-15, 2023.

[54] H. Ma, Y. Zhang, N. Thuerey, X. Hu, and O. J. Haidn, "Physics-driven learning of the steady Navier-Stokes equations using deep convolutional neural networks," *arXiv preprint arXiv:2106.09301,* 2021.

[55] M. D. Ribeiro, A. Rehman, S. Ahmed, and A. Dengel, "DeepCFD: Efficient steady-state laminar flow approximation with deep convolutional neural networks," *arXiv preprint arXiv:2004.08826,* 2020.

[56] J. Schulman, F. Wolski, P. Dhariwal, A. Radford, and O. Klimov, "Proximal policy optimization algorithms," *arXiv preprint arXiv:1707.06347,* 2017.

[57] S. Hochreiter and J. Schmidhuber, "Long short-term memory," *Neural computation,* vol. 9, no. 8, pp. 1735-1780, 1997.

[58] H. Jasak, A. Jemcov, and Z. Tukovic, "OpenFOAM: A C++ library for complex physics simulations," in *International workshop on coupled methods in numerical dynamics*, 2007, vol. 1000, pp. 1-20.

[59] L. Alzubaidi *et al.*, "Review of deep learning: Concepts, CNN architectures, challenges, applications, future directions," *Journal of big Data,* vol. 8, pp. 1-74, 2021.

[60] M. Tschannen, O. Bachem, and M. Lucic, "Recent advances in autoencoder-based representation learning," *arXiv preprint arXiv:1812.05069,* 2018.

[61] A. Mousavian, D. Anguelov, J. Flynn, and J. Kosecka, "3d bounding box estimation using deep learning and geometry," in *Proceedings of the IEEE conference on Computer Vision and Pattern Recognition*, 2017, pp. 7074-7082.

[62]	M. Yousef, F. Iqbal, and M. Hussain, "Drone forensics: A detailed analysis of emerging DJI models," in *2020 11th International Conference on Information and Communication Systems (ICICS)*, 2020: IEEE, pp. 066-071.

[63]	Y. Bengio, J. Louradour, R. Collobert, and J. Weston, "Curriculum learning," in *Proceedings of the 26th annual international conference on machine learning*, 2009, pp. 41-48.